\def\year{2019}\relax
\pdfoutput=1
\documentclass[letterpaper]{article} 
\usepackage{aaai19}  
\usepackage{times}  
\usepackage{helvet}  
\usepackage{courier}  
\usepackage{url}  
\usepackage{graphicx}  
\usepackage{subfig}
\usepackage{alphalph}
\usepackage{calc}
\usepackage{enumitem}  

\usepackage{booktabs}
\usepackage{float}
\usepackage{array,tabularx}
\usepackage{amsmath}
\usepackage{amsfonts}

\newcommand{\argmax}[1]{\underset{#1}{\operatorname{argmax} \hspace{0.05cm}}}

\begin{document}
%
\title{Deterministic Implementations for Reproducibility in Deep Reinforcement Learning}

\author{Prabhat Nagarajan\thanks{This work took place primarily while this author was a student at the University of Texas at Austin.}\\ 
Preferred Networks\\
Tokyo, Japan\\
prabhat@preferred.jp
\And Garrett Warnell\\ 
Computational and Information Sciences Directorate\\
U.S. Army Research Laboratory\\
garrett.a.warnell.civ@mail.mil
\AND Peter Stone\\
Department of Computer Science\\
The University of Texas at Austin\\
pstone@cs.utexas.edu}

\maketitle
\begin{abstract}
While deep reinforcement learning (DRL) has led to numerous successes in recent years, reproducing these successes can be extremely challenging. One reproducibility challenge particularly relevant to DRL is nondeterminism in the training process, which can substantially affect the results. Motivated by this challenge, we study the positive impacts of deterministic implementations in eliminating nondeterminism in training. To do so, we consider the particular case of the deep Q-learning algorithm, for which we produce a deterministic implementation by identifying and controlling all sources of nondeterminism in the training process. One by one, we then allow individual sources of nondeterminism to affect our otherwise deterministic implementation, and measure the impact of each source on the variance in performance. We find that individual sources of nondeterminism can substantially impact the performance of agent, illustrating the benefits of deterministic implementations. In addition, we also discuss the important role of deterministic implementations in achieving exact replicability of results.
\end{abstract}

\section{Introduction} \label{intro}
The reproducibility of algorithms and results in deep reinforcement learning (DRL) is paramount. For one, the reproducibility and verifiability of results ensure that DRL is on stable footing. Furthermore, reproducibility increases the rate at which DRL research progresses, since people can more easily build off of prior work. DRL in particular has witnessed several remarkable results that have elevated the field of artificial intelligence. These results include one of the first systems to play video games while learning directly from pixels \cite{dqn}, algorithms enabling agents to perform complex locomotion behaviors \cite{locomotivebehaviors}, and a system that plays the game of Go at a superhuman level \cite{alphago}. Along with these impressive successes, however, has come both an increased difficulty and increased need to reproduce successful DRL algorithms \cite{deeprlmatters}. In this paper, we argue that deterministic implementations are crucial in tackling reproducibility challenges faced by DRL.

Perhaps the most obvious reproducibility goal that deterministic implementations help achieve is \textit{replicability}. There is a distinction between the general goal of \textit{reproducibility} and the stricter notion of \textit{replicability}. We define them here as follows:

\begin{description}[leftmargin=!,labelwidth=\widthof{\bfseries Reproducibility}]
\item[Reproducibility:] the ability of an experiment to be repeated with minor differences from the original experiment, while achieving the same qualitative results. 
\item[Replicability:] the ability of an experiment to be repeated exactly, producing the same quantitative results.
\end{description}

\noindent In DRL, it is possible that an empirical result may be a false positive. In such a scenario, an independent party may not be able to reproduce the result in question. Rather than misattributing the inability to reproduce an experiment to ghost factors, replicability enables experiments to be repeated exactly and investigated further. This can be done regardless of the truth of the result, and therein lies its benefit. In the context of DRL, achieving replicability requires a \textit{deterministic implementation} to be run under identical \textit{experimental conditions}. \textit{Experimental conditions} refer to the software and hardware conditions under which a computational experiment is executed. A \textit{deterministic implementation} is defined as:

\begin{description}[style = multiline, leftmargin=!,labelwidth=\widthof{\bfseries implementation:}]
\item[Deterministic implementation:] a computer program that, when run under some fixed experimental conditions, will always produce identical outputs for a given input. 
\end{description}

\noindent Note that deterministic implementations alone are not replicable experiments. A replicable experiment consists of a deterministic implementation \textit{and} fixed experimental conditions. If a deterministic implementation is executed on different hardware or is compiled differently from the original experiment, the experimental results may not be replicated.

Another related way in which deterministic implementations can contribute to reproducibility goals is through their ability to eliminate nuisance noise from results, which can benefit statistical testing. Even with access to the original implementation, nondeterminism in the training process can cause large variation in results, making them difficult to reproduce. DRL is uniquely susceptible to nondeterminism in that the agent learns from a nonstationary distribution of experiences, which in turn is influenced by nondeterministic environments and nondeterministic policies. Imagine how a small difference between two agents' early experiences can proliferate throughout the training process as their nonstationary experience distributions drift apart. One small difference can result in drastically different outcomes for the two agents. It is this \textit{cascading effect} that makes DRL particularly susceptible to nondeterminism. This susceptibility is the motivation for our sensitivity analysis, in which we measure the effect of individual sources of nondeterminism on the variance in performance throughout training.

In this paper, our primary contributions are to:
\begin{enumerate}
    \item identify the sources of nondeterminism in DRL and describe a deterministic implementation of deep Q-learning (which we make publicly available), and
    \item measure the sensitivity of an agent's performance to individual sources of nondeterminism.
\end{enumerate}
Given that deterministic implementations are a prerequisite to the goal of replicability, we also identify some experimental conditions that form the gap between deterministic implementations and replicability.

\section{Background}
We now provide a brief background of the Markov decision process formulation of reinforcement learning problems and of the deep Q-learning algorithm, which is our algorithm of interest in this paper.
\subsection{Markov Decision Processes}
Reinforcement learning (RL) problems are formulated as \textit{Markov decision processes} (MDPs). An MDP is a tuple $(\mathcal{S}, \mathcal{A}, P, \gamma, \mathcal{R})$, where $\mathcal{S}$ denotes the set of \textit{states} within the environment and $\mathcal{A}$ denotes the set of actions available to the agent within the environment. The agent acts at discrete timesteps where, at each timestep, it observes a state, performs an action, and transitions to another state.
This is formalized by the \textit{transition model} $P$, where $P(s^{\prime} | s, a)$ is the probability that the agent transitions to state $s^{\prime}$ when performing action $a$ in state $s$.
The \textit{discount rate} $\gamma \in [0,1]$ specifies the agent's preference for immediate rewards versus future rewards.
The \textit{reward function} $\mathcal{R} : \mathcal{S} \times \mathcal{A} \times \mathcal{S} \rightarrow \mathbb{R}$ provides the agent with reward $\mathcal{R}(s,a,s^{\prime})$ as it transitions to state $s^{\prime}$ after performing action $a$ in state $s$.

Given an MDP, an RL agent's objective is to learn a policy $\pi : \mathcal{S} \times \mathcal{A} \rightarrow [0,1]$ mapping a state-action pair $(s,a)$ to the probability that the agent performs action $a$ in state $s$. Specifically, the agent tries to learn an \textit{optimal policy} $\pi^{*}$, a policy that maximizes the agent's expected cumulative discounted reward $\mathbb{E}[\sum_{t=0}^{\infty} \gamma^{t}\mathcal{R}(s_{t}, a_{t}, s_{t+1})].$ Often, rather than directly learning an optimal policy $\pi^{*}$, the agent learns the \textit{optimal state-action value function} $Q^{*} : \mathcal{S} \times \mathcal{A} \rightarrow \mathbb{R}$, which maps a state-action pair $(s,a)$ to the expected cumulative discounted reward the agent receives if action $a$ is taken in state $s$ and optimal actions are performed thereafter. If the agent learns $Q^{*}$, then an optimal policy can be to perform action $\argmax{a} Q^{*}(s,a)$ in state $s$.

\subsection{Deep Q-learning}
Deep Q-learning \cite{dqn2013,dqn} is an algorithm that trains a deep neural network through Q-learning \cite{qlearning} to approximate the optimal state-action value function $Q^{*}$ in high dimensional state spaces. The algorithm is applied to the Arcade Learning Environment (ALE) \cite{ale}, an evaluation platform that provides RL agents with an interface to play Atari games. Deep Q-learning is able to achieve human-level performance in many of these games while learning directly from pixel representations of the state (i.e. game frames).  As the agent interacts with its environment, it maintains a \textit{replay buffer} $\mathcal{D}$ of its last $N$ transitions (typically $N=1$ million). Each entry in this replay buffer contains a tuple $(s_{t}, a_{t}, r_{t}, s_{t+1})$, representing the state, action, reward, and subsequent state, respectively. The network representing the state-action value function being learned is termed a deep Q-network (DQN), where $Q(s,a;\boldsymbol{\theta})$ represents the predicted state-action value under the DQN parameters $\boldsymbol{\theta}$. The algorithm also maintains a separate \textit{target network} $Q(s,a; \boldsymbol{\theta}^{-})$, where $\boldsymbol{\theta}^{-}$ represents the parameters of a network from a prior training iteration. Periodically, the target network parameters are reset to equal the DQN parameters: $\boldsymbol{\theta}^{-} \leftarrow \boldsymbol{\theta}$. To train the DQN at iteration $i$, the agent minimizes the loss \cite{dqn}:

\begin{align*}
\mathbb{E}_{(s, a, r, s^{\prime}) \sim \mathcal{U}(\mathcal{D})}\bigg[ \big( r + \gamma \max_{a^{\prime}}Q(s^{\prime},a^{\prime}; \boldsymbol{\theta}_{i}^{-}) - Q(s,a; \boldsymbol{\theta}_{i}) \big)^{2} \bigg].
\end{align*}

 Following techniques from stochastic optimization, the agent randomly samples minibatches of transitions uniformly from the replay buffer, uses the DQN and the target network to compute the loss, and then updates the DQN's weights.

\section{Deterministic Implementations}
In order to produce a deterministic implementation of deep Q-learning, we must first identify all the sources of nondeterminism that are present in its implementation. Once identified, controlling or eliminating these sources from the learning process is sufficient for obtaining a deterministic implementation.

\subsubsection{Sources of Nondeterminism}
While the exact sources of nondeterminism depend on the algorithm, problem domain, libraries, etc., we identify those sources common to most DRL algorithm implementations.

\begin{itemize}
    \item \textbf{GPU} Neural networks are typically trained on graphics processing units (GPUs). Many numerical operations performed on the GPU are nondeterministic by default.
    \item \textbf{Environment} The environment in reinforcement learning can be stochastic. That is, the transitions can be random.
    \item \textbf{Policy} During training, reinforcement learning agents typically employ a stochastic policy. That is, the agent's action is drawn from a non-degenerate distribution over the available actions.
    \item \textbf{Network initialization} Prior to training, the weights of the neural network are randomly initialized.
    \item \textbf{Minibatch sampling} When training neural networks, several algorithms sample random minibatches of training data from some dataset.
\end{itemize}

All of these sources of nondeterminism are present in deep Q-learning. For example, the deep Q-network is randomly initialized and trained on a GPU. The agent employs an $\epsilon$-greedy stochastic policy during training, whereby at each timestep, it performs a random action with probability $\epsilon$. Deep Q-learning uses random minibatch sampling from the replay buffer when training the neural network. The algorithm was originally developed for the first version of the ALE \cite{ale}, where the environment is completely deterministic. However, in the most recent version of the ALE \cite{revisitingale}, a new form of stochasticity is added to the environment, in the form of ``sticky actions''. In an environment with sticky actions, an agent's \textit{previous} action is performed in the environment with probability $p=0.25$, and the agent's \textit{chosen} action is performed with probability $1-p$. The last form of nondeterminism in deep Q-learning, not listed above, is the use of ``no-op'' or do-nothing actions. In the deterministic ALE ($p=0.0$) for which deep Q-learning was originally designed, the agent performs a random  number of no-op actions at the beginning of each episode in order to randomize the initial state within the deterministic environment.

\subsubsection{Implementation: Eliminating Nondeterminism} \label{detimpl}
Our implementation of deep Q-learning is written in Python using the PyTorch library \cite{pytorch}. PyTorch\footnote{We selected PyTorch for its ease of controlling GPU nondeterminism. However, the sources of nondeterminism can depend on the deep learning library used. For example, some versions of Tensorflow have certain nondeterministic functions that require workarounds and enabling GPU determinism is not straightforward. In fact, we were not able to do so.} exposes a modifiable boolean variable that allows us to enable or disable deterministic numerical computations on the GPU. Furthermore, PyTorch permits us to set the seed used to initialize the weights of the deep Q-network, allowing us to obtain identical initial networks on separate runs. To control for no-ops, exploration, and minibatch sampling, we assign each of these sources of nondeterminism its own seeded random number generator. Any random operations required for no-ops, exploration, or minibatch sampling are then implemented using the assigned random number generator. Thus, across training runs, the same ``random'' number of no-ops are performed at the beginning of episodes. Exploratory actions are identical and occur at consistent timesteps across runs. Similarly, the same minibatch indices are sampled across runs. In controlling environment nondeterminism, there are two possibilities. The first is when we use a deterministic environment, where we simply set $p=0.0$. The second scenario is when we wish to use the stochastic ALE. In this scenario, we set $p=0.25$, and introduce a ``sticky action'' seed, which we use to create a random number generator to implement sticky actions. That is, we use this random number generator to decide whether to repeat the previous action or perform the new one. 

If all experimental conditions are held fixed (as is the case in our experiments), and all sources of nondeterminism are controlled in this fashion, we then achieve identical results on separate training runs, as desired. To validate the equivalence of separate runs, we verify that the learned weights of the neural network are identical at intervals throughout training.  

We use the standard DQN architecture \cite{dqn} for all agents. Each agent's policy during training is an $\epsilon$-greedy policy, where at each timestep, the agent either performs a random action with probability $\epsilon$, or the greedy action $\argmax{a} Q^{*}(s,a; \boldsymbol{\theta})$ with probability $1-\epsilon$. The value $\epsilon$ is initially set to $1.0$ and is linearly annealed to $0.1$ over the first million frames, remaining at $0.1$ thereafter. For agents trained in a stochastic environment, we anneal $\epsilon$ to 0.01 over the first million frames, after which it remains at 0.01. We make our deterministic implementation\footnote{\url{https://github.com/prabhatnagarajan/repro_dqn}} publicly available.

\subsubsection{Experimental Conditions and Replicability}

Recall that deterministic implementations are necessary but not sufficient for achieving replicability. Varying the experimental conditions can cause a deterministic implementation to produce different results\footnote{In the Supplemental Material, we plot the learning curves of a deterministic implementation executed on two separate machines, producing different curves.}. While we do not aim to identify all hardware or software conditions that can influence replicability, it is useful to be aware of some experimental conditions that can impede replicability.

On the software side, the deep learning library version can influence replicability. For example, some versions of TensorFlow \cite{tf} have library functions that are nondeterministic. Furthermore, in some scenarios, the library functions must be run as single-threaded in order to achieve determinism. Regarding GPU-related software, according to the cuDNN documentation (cuDNN underlies many deep learning libraries), bit-wise reproducibility cannot be ensured, since implementations for some routines vary across versions \cite{cudnndocs}. From the hardware side, running the same deterministic implementation on a CPU can yield different results from running deterministically on a GPU. This can be due to several reasons \cite{cpuvgpu}, including differences in available operations and in the precision between the CPU and GPU. Further, when a deterministic implementation is run on two different GPU architectures, it may produce different results, since code generated by the compiler is then compiled at run-time for a specific target GPU \cite{cudafaq,cudnndocs}.

\section{Experiments}
In order to quantify the benefit of controlling nondeterminism, we use our deterministic implementation to systematically allow individual sources of nondeterminism to influence the training process. We measure the sensitivity of the agent's performance to each individual source of nondeterminism. Specifically, our measure of sensitivity is the standard deviation in the agent's achieved performance (reward). Our sensitivity analysis highlights the benefit of deterministic implementations. By permitting just a single source of nondeterminism to influence the training process, we demonstrate how impactful even a single source of nondeterminism is on the variability of performance.

\subsection{Sensitivity Analysis}
\subsubsection{Training}

In our experiments, we train six groups of networks using our deterministic implementation. Since we are interested in the reproducibility of published results, we emulate practical experimental scenarios by training five networks for each experimental group, since a sample size of five is commonly used in DRL papers \cite{revisitingale,gorila}.

 The first group, which we call the ``deterministic'' group, consists of five networks trained with the same random seeds and with deterministic GPU operations enabled across all runs. The ``GPU'' group has all settings identical to the deterministic group except with nondeterministic GPU operations enabled. The ``environment'' group is trained in a stochastic environment, with $p=0.25$, whereas all other groups are trained in a deterministic environment ($p=0.0$). Each agent in the environment group has a different sticky action seed to inject environment nondeterminism, ensuring each run is different. Except for setting $p=0.25$ and using a different sticky action seed for every run, all other settings of environment group are identical to the deterministic group's settings. The ``exploration'' group consists of five networks trained with different random exploration seeds and all other settings identical to the deterministic group. The ``initialization'' group consists of five networks each trained with a different set of randomly initialized weights, with all other settings identical to the deterministic group. The ``minibatch'' group is also trained with identical settings to the deterministic group, except each of the five agents has a different minibatch seed.

All of our agents are trained on the Atari game \textsc{Breakout}, a domain where the agent uses a paddle to hit a moving ball while trying to eliminate rows of bricks from the game screen. We choose this domain due to its widespread popularity in the DRL research community. All of our agents were trained for 20 million time
steps in the ALE and are evaluated after every 250K timesteps of training. At each of these evaluation intervals, we measure the mean and standard deviation of the performances of the agents within a group. The hardware and software conditions are held constant for all experiments (see Supplemental Material for experimental conditions).
 
\subsubsection{Evaluation Protocol}
When we measure the performance of our agents, we want to ensure that any differences in performance are a result of differences between their Q-networks. In doing so, we ensure that we are measuring performance differences due to an individual source of nondeterminism, since the differences between agents' trained Q-networks are solely due to a source of nondeterminism influencing the learning process. As such, we evaluate the agents over 100 episodes (each episode is capped at five minutes of play) using a greedy policy, so that any deviations between agents' policies are a consequence of their different Q-networks. However, in the deterministic ALE ($p=0$), repeating a greedy policy 100 times results in 100 identical trajectories. Therefore, to ensure that our evaluation protocol is comprehensive and measures an agent's performance in a diverse set of conditions, we have each of our 100 episodes begin with a unique start state, from which an agent performs a greedy policy. This protocol contrasts from the typical ways of introducing diversity in the evaluation, which involve injecting some stochasticity into the policy. However such evaluation protocols can confound our results, as we discuss in the Supplemental Material. To produce a unique start state for each episode, we begin each episode with a predetermined action sequence consisting of dozens of actions, with each sequence ending in a unique state. Though the start states are generated randomly for diversity, we still ensure that the start states are not poor states that impede an agent's ability to perform well in an evaluation episode. The full details of the start state generation are in the Supplemental Material.

Since the environment group is trained with $p=0.25$, this group uses a slightly different evaluation protocol. We still have each agent performing a greedy policy for 100 episodes, where each episode is capped at five minutes. However, rather than beginning episodes at unique start states, each episode uses a unique sticky action seed to implement stochasticity for that episode. The 100 sticky action seeds used for the evaluation episodes are held constant across all evaluations of the environment group.  This protocol again ensures that we have 100 different episodes in which deviations between two agents' trajectories are caused solely by disparities in their decisions made within that episode.

\subsubsection{Results}

\captionsetup[subfigure]{subrefformat=simple,labelformat=simple,listofformat=subsimple}
\renewcommand\thesubfigure{(\alph{subfigure})}
\begin{figure*}[ht]
\begin{center}
	\subfloat[][Deterministic]{ \label{detscore}
    	\includegraphics[width=0.28\linewidth]{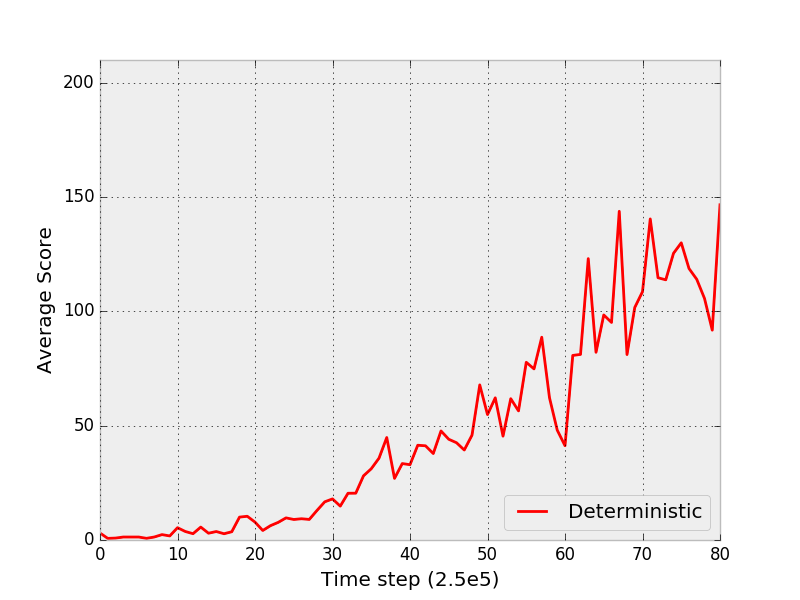}}
    \subfloat[][GPU]{ \label{gpuscore}
    	\includegraphics[width=0.28\linewidth]{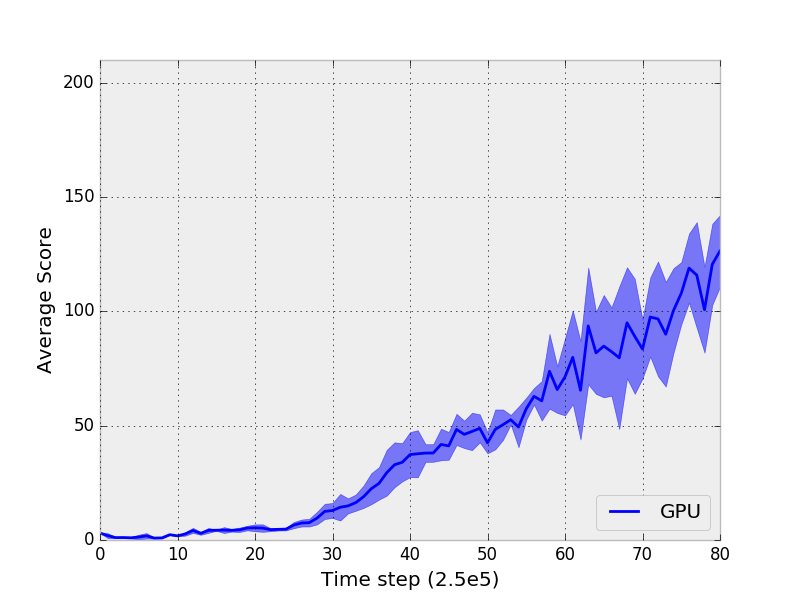}}
	\subfloat[][Environment]{ \label{envscore}
    	\includegraphics[width=0.28\linewidth]{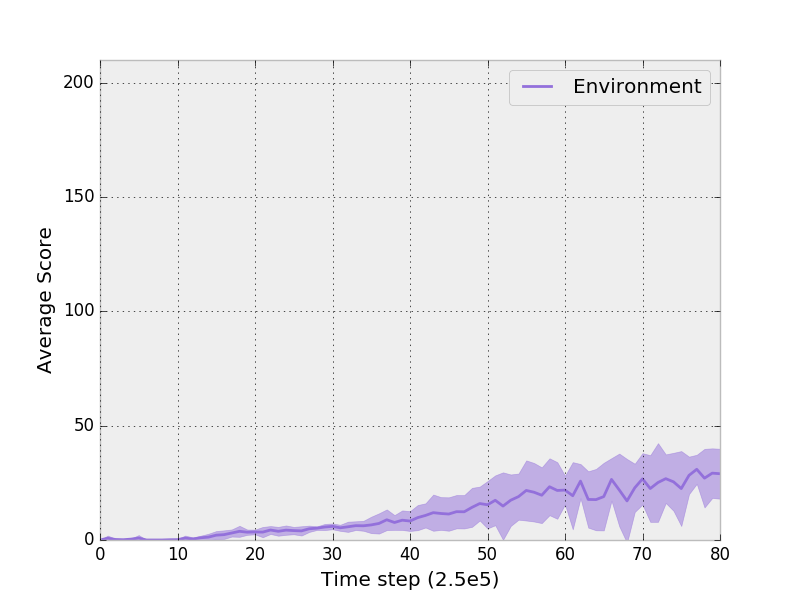}} \\
    \subfloat[][Exploration]{ \label{expscore}
    	\includegraphics[width=0.28\linewidth]{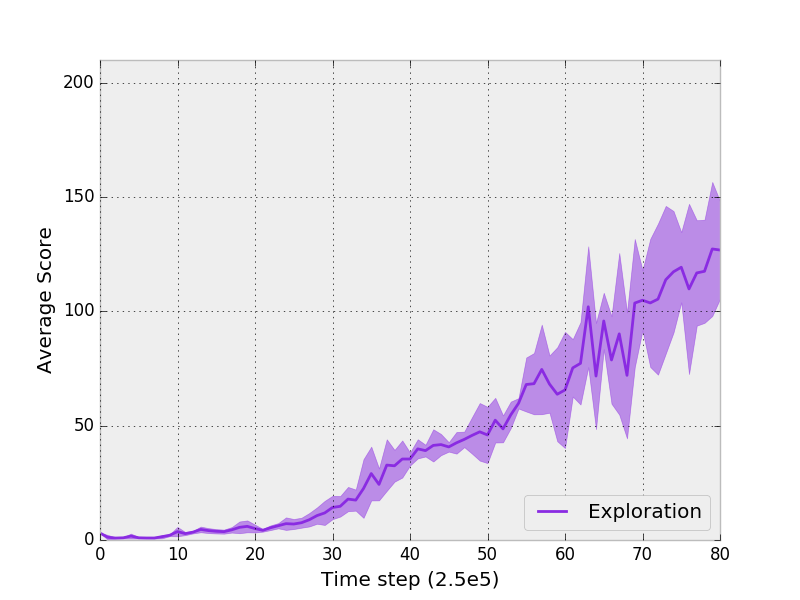}}
    \subfloat[][Initialization]{ \label{initscore}
    	\includegraphics[width=0.28\linewidth]{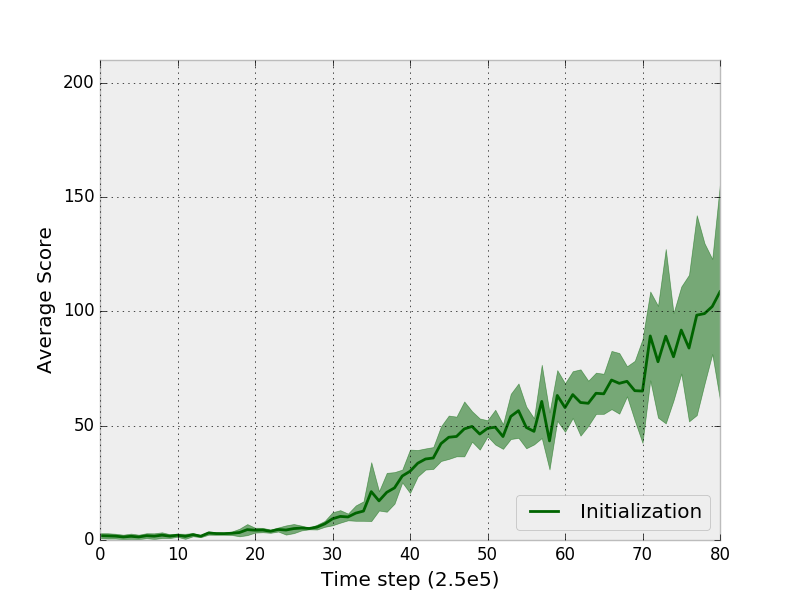}}
    \subfloat[][Minibatch]{ \label{miniscore}
    	\includegraphics[width=0.28\linewidth]{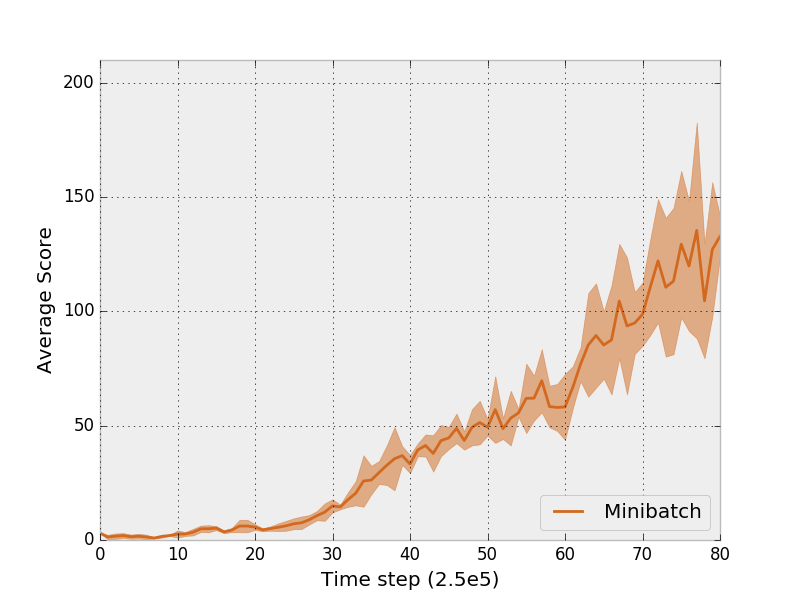}} 
\end{center}
\caption{The game scores for our six experimental groups. Solid curves depict the mean score. Shaded areas represent values within one standard deviation of the mean score. The absence of a shaded area indicates identical results across all five runs.}
\label{performances}
\end{figure*}

\begin{table*}[ht] \centering
\begin{small}
\begin{tabular}{@{}l|cccccc@{}}\toprule
 \textbf{Metric} \hfill & \textbf{Deterministic} & \textbf{GPU} & \textbf{Environment} & \textbf{Exploration} & \textbf{Initialization} & \textbf{Minibatch}\\ \midrule
\textbf{Average Score} (\textit{Best}) & 146.7 & 141.9 & 33.6 & 148.6 & 131.2 & 153.38 \\ 
\textbf{Standard Deviation} (\textit{Best}) & 0.0 & 8.8 & 8.7 & 17.0 & 31.0 & 32.96 \\ 
\textbf{Relative Standard Deviation} (\textit{Best}) & 0.0\% & 6.22 \% & 25.96\% & 11.42\% & 23.61\% & 21.49\% \\ 
\midrule
\textbf{Average Score} (\textit{Final}) & 146.7 & 126.5 & 29.0 & 126.9 & 108.6 & 132.84 \\ 
\textbf{Standard Deviation}\newline (\textit{Final}) & 0.0 & 15.7 & 10.9 & 21.4 & 47.4 & 8.89 \\ 
\textbf{Relative Standard Deviation} (\textit{Final}) & 0.0\% & 12.41\% & 37.65\% & 16.85\% & 43.61\% & 6.69\% \\ 
\bottomrule
\end{tabular}
\end{small}
\caption{The mean, standard deviation, and relative standard deviation of scores in \textsc{Breakout} for six experimental groups.}
\label{perfresultstable}
\end{table*}

The results for our sensitivity analysis are depicted in Figure \ref{performances} and in Table \ref{perfresultstable}. We show six graphs, one per experimental group, plotting the mean performance of the agents within the group (where performance is the mean score on 100 episodes), with the shaded areas depicting values within one standard deviation of the mean performance. In Table \ref{perfresultstable}, we show the mean and standard deviation in performance for the agents within each group at the end of training, i.e., after 20 million timesteps. We also depict the mean and standard deviation of the \textit{best-scoring} networks within a group. The best scoring network of a single training run refers to the network parameters that achieve the highest performance across all evaluation intervals during a single training run. It is common practice in DRL \cite{revisitingale} to report the mean score of the best-scoring networks of several training runs and thus we do so here, to better reflect the impact of nondeterminism on reported performance in practice. We also report the relative standard deviation in performance, in order to provide a domain-agnostic measure of variance, since the numerical score is specific to \textsc{Breakout}.

\section{Discussion}

Consider Figure \ref{detscore}, which depicts the learning curve of the deterministic group. The key takeaway from this figure is that there is no shaded area. All five curves within the deterministic group overlap exactly, with zero variance at every point of the learning curve, a direct consequence of deterministic training.

\begin{figure}[ht]
    \centering
    \includegraphics[width=0.3\textwidth]{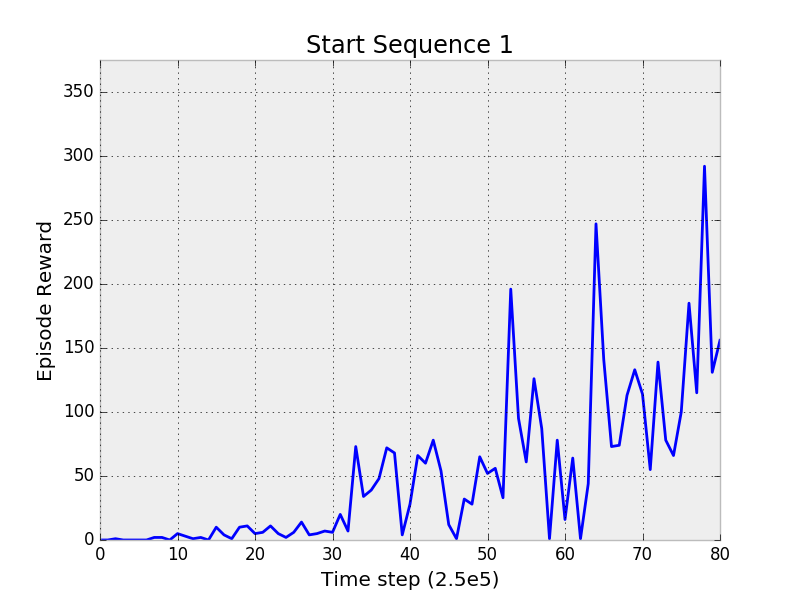}
    \caption{The score of the deterministic agent on an individual start state.}
    \label{indiecurve}
\end{figure}

Notice, however, how the deterministic learning curve is less stable than the other curves, despite representing an average of 100 episodes. This instability can be observed more dramatically when we consider the performance of the agent on an individual start state, as in Figure \ref{indiecurve}. We observe sharp fluctuations in performance even between consecutive evaluation intervals, demonstrating that DQNs are very sensitive to minor changes in weights, a known phenomenon \cite{dqn2013}. Note that we observe results similar to Figure \ref{indiecurve} for all 100 start states (see Supplemental Material). We know that deterministic implementations are needed to replicate curves exactly, but Figure \ref{indiecurve} shows that without deterministic implementations, it is unlikely that we can even closely reproduce performance for individual start states.

The remainder of the curves in Figure \ref{performances} all exhibit a key feature: growing variance as learning progresses. For all of these curves, we observe low variance early in training, likely due to the controlled sources of nondeterminism which keep the performances similar. However, as the agents learn and their experiences and network weights diverge, we observe the variance in performance growing larger. This growing variance is a characteristic of the cascading effect, where small differences influence the policy, thereby producing more differences and larger variance in performance. The GPU curve, Figure \ref{gpuscore}, particularly underscores the cascading effect. The GPU is the only source of nondeterminism that exists \textit{outside} of the algorithm itself. In theory, if we had infinite precision, the GPU curve would have no variance. However, small differences in computation compound upon one another as training progresses, resulting in the 12.41\% relative standard deviation at the end of training, as shown in the GPU column of Table \ref{perfresultstable}. We often ignore GPU nondeterminism as negligible when reproducing research. However, GPU nondeterminism epitomizes the benefit of deterministic implementations, as it shows that even if all implementations details and all other sources of nondeterminism are held fixed, small errors can compound throughout training and cause noise in the produced results.

When considering the results for each individual source of nondeterminism, it is important to reflect on the scenarios in which eliminating nondeterminism reduces variance. We know that for any of our nondeterministic experimental groups, differences between networks will inevitably arise, after which the cascading effect can inflate these differences. However, we can still benefit from controlling some sources of nondeterminism, since controlling nondeterminism can ameliorate the amount by which initial differences arise. This is reflected in Table \ref{perfresultstable}. For example, the GPU results have the lowest relative standard deviation of the best networks. The members of the GPU group start with identical initial networks and share identical random exploration seeds in a deterministic environment. Consequently, the networks of the GPU group have identical experiences for tens of thousands of time steps, and populate their replay buffers with similar experiences. Further, it takes tens of thousands of steps before the nondeterministic GPU operations create substantial enough differences to manifest as different action selections by the agents. For all of these reasons, the differences caused by GPU nondeterminism are minimal early in training, relative to other sources of nondeterminism, before the cascading effect plays a significant role.

In contrast, consider the environment group, where eliminating nondeterminism is the least beneficial in terms of reducing the relative standard deviation in the best networks' performance. The networks in this group operate in a stochastic environment, so that virtually instantly the shared exploration and minibatch seeds lose their benefit. The stochastic environments immediately cause agents to have different experiences, and consequently their replay buffers are different. Further, the exploration seed causes identical exploratory actions to be chosen at the same time steps, but this is ineffective for the environment group, since the agents quickly go to different states due to the stochasticity. The only source of nondeterminism that may have an enduring benefit on reducing variance is the network initialization, which alone proves insufficient for majorly reducing variance.

Unsurprisingly, the initialization group has the highest relative variance in performance amongst both its final networks and its best networks of any experimental group trained in the deterministic version of \textsc{Breakout}. This is expected, since varying the network initialization induces a permanent difference between the agents that is reflected throughout training. Having $\epsilon$ set to $1.0$ at the beginning of training causes the initialization agents to share many early experiences. However, it appears that these experiences are insufficient to overcome the permanently ingrained differences of the network initializations themselves. 

The exploration group has less variance than the initialization group, though more than the GPU group. The exploration group's networks share random initializations but their experiences and replay buffers desynchronize immediately due to the high exploration early in training. Though this group exhibits less variance than the initialization group, we did not find a statistically significant difference between their variances. Therefore we are unable to conclude that the effect of random initializations on the variance in performance is greater than that of having different early experiences.

Contrary to what we might expect, minibatch sampling has the highest variance in best performance (though not relative variance). This is counterintuitive, since the large replay buffer size (1 million) makes the experience distribution, from which we sample minibatches, more stationary. A closer look at the data indicates the cause of this large variance is an outlier. Notice how the average performance is comparatively higher for the minibatch group's best networks than the other groups. This is due to one network scoring over 200 at its best. Unfortunately, as we see in Figure \ref{indiecurve}, the agent's performance on individual start states is volatile and reporting the performance of the best network can be quite noisy and susceptible to spikes in the learning curve. This is an inherent drawback to this form of reporting performance, and in fact it has been recommended \cite{revisitingale} that future ALE research avoid this form of reporting.

Our sensitivity analysis demonstrates several key points that highlight the benefit of deterministic implementations. Perhaps the most important observation is that allowing some sources of nondeterminism to remain uncontrolled can result in large variance, as shown by the initialization group. It is scenarios like these that we aim to avoid with deterministic implementations. Another obvious, yet important observation is that deterministic runs produce no variance in the learning curve or reported performance, as intended. In all nondeterministic groups, we observe a growing variance in performance, with the cascading effect likely playing a major role. We also observe that some sources of nondeterminism produce much less variance than others. We even find that for the final networks, at a significance level of $\alpha=0.1$, the initialization group has statistically different variance from the GPU group. This is noteworthy because it shows that controlling even \textit{some} sources of nondeterminism can reduce variance. This is particularly promising for real-world domains such as robotics, where it may be impossible to have deterministic implementations. Lastly, it is very important to take note that we have demonstrated the impact of nondeterminism using a deterministic evaluation protocol, where performance differences are all attributable to the policies of the agents. We may find that stochastic evaluation protocols increase the variance we observe for individual sources of nondeterminism. 

\section{Related Work}
In this section, we discuss prior work on replicability, reproducibility in DRL, as well as related research on the ALE.

\subsection{Replicability and Experimental Conditions}
There have been studies to identify and examine the experimental conditions that can affect the outcome of computational experiments. For example, Gronenschild et al. \cite{experimentalconditions} analyze the results produced by a software package called FreeSurfer \cite{freesurfer}, which is used to make measurements in studies of neuroanatomical structures. They test measurements produced while using FreeSurfer under different experimental conditions. They vary the FreeSurfer version, the operating system, and the workstation (hardware), and find that results can vary significantly when experimental conditions are changed. 
\subsection{Reproducibility Efforts}
There have been several recent efforts to address reproducibility through the public release of environments and implementations \cite{baselines,gym,ale,dopamine,deepmindlab,deepmindcontrolsuite}, enabling better benchmarking, experimentation, and comparison of DRL algorithms. We also see a rise in the development of reproducibility-friendly software. For example, AWS Docker containers \cite{awsdocker} and CodaLab Worksheets \cite{codalab} can be used to achieve replicability by packaging the experimental conditions with the code. Sumatra \cite{sumatra} is a tool that can be used to control dependencies and assist with version control in reproducible research. CDE (Code, Data, Environment) \cite{cde} packages software dependencies that are needed to rerun Linux-based experiments on other machines. Lastly, Jupyter notebooks \cite{jupyter} enable researchers to readily have code with explanations packaged together. Further, these notebooks can run inside containers (e.g. Docker), permitting replicability.

\subsection{Reproducibility in Deep Reinforcement Learning}
While reproducibility has been explored across artificial intelligence and machine learning \cite{replnotrepro,reproai}, reproducibility in DRL remains relatively uncharted. In the context of DRL, the effects of hyperparameters, codebases, evaluation metrics, random seeds, and aspects of the environment have been studied to a degree \cite{deeprlmatters,reprobench,revisitingale}. Henderson et al. \cite{deeprlmatters} show that the choice of hyperparameters, network architecture, reward scale, and random seeds can have a dramatic effect on the performance of an agent. They show that some algorithms perform better than others in environments with stable dynamics, while performing worse in environments with unstable dynamics. They also find that differences in implementation details between codebases implementing the same DRL algorithm can result in drastically different performances. Perhaps their most shocking result is that two groups of networks trained from the same algorithm implementation can yield statistically significantly different performances solely due to differences in global random seeds.

Our paper differs from prior research in that we investigated individual sources of nondeterminism in isolation, as opposed to the aggregate effect of random seeds. Furthermore, our study of environment stochasticity studied the effect of injecting stochasticity into a deterministic environment, as opposed to comparing stochastic tasks to deterministic tasks when the  underlying task semantics are inherently different. Finally, prior work focuses on different DRL algorithms and are concerned more broadly with reproducibility, as opposed to our focus on deterministic implementations, which addresses both replicability and reproducibility.

Reproducibility in DRL has also been examined from the perspective of statistical hypothesis testing by Colas et. al \cite{howmanyseeds}. They describe good statistical practices for algorithm comparison as well the selection of an appropriate sample size in DRL. 

\subsection{The Arcade Learning Environment}
While not necessarily in the domain of reproducibility, there have been studies on the impact of deterministic environments in the ALE \cite{determinismale,revisitingale}. These studies find that the deterministic environments in the ALE can be exploited by naive agents that can perform well simply by memorizing action sequences. To combat this naive way of achieving success, these studies examine several methods of injecting stochastity into the environment, such as no-ops, exploration, and sticky actions. It should be noted that these studies are performed in the presence of other forms of nondeterminism (e.g. GPU nondeterminism), unlike our experiments.

\section{Conclusion}
In this paper, we explored the important role of deterministic implementations for achieving reproducibility in DRL. We identified the various sources of nondeterminism at play, and described how to produce a deterministic implementation of deep Q-learning. We have discussed the relationship and distinction between replicability and deterministic implementations, and noted some of the factors that inhibit replicability. Our sensitivity analysis on sources of nondeterminism demonstrates the large variance in results that can occur due to individual sources of nondeterminism alone, further supporting the need for deterministic implementations.

Given the many benefits of deterministic implementations, we encourage the research community to embrace them in all aspects of research. We hope to see wider use and dissemination of deterministic implementations and replicable experiments in the future.

\section{ Acknowledgments}
The authors would like to thank Naren Manoj, Brahma Pavse, Darshan Thaker, and Ewin Tang for their useful comments, suggestions, and reviews on prior versions of this work.

 This work has taken place in the Learning Agents Research Group (LARG)
 at UT Austin.  LARG research is supported in part by NSF (IIS-1637736,
 IIS-1651089, IIS-1724157), Intel, Raytheon, and Lockheed Martin.  Peter
 Stone serves on the Board of Directors of Cogitai, Inc.  The terms of
 this arrangement have been reviewed and approved by the University of
 Texas at Austin in accordance with its policy on objectivity in
 research. 

\bibliographystyle{aaai}

\onecolumn 
\begin{center}
\textbf{\LARGE Supplemental Material}
\end{center}

\section{Replicability vs. Determinism}

\begin{figure*}[h]
    \centering
    \includegraphics[width=0.5\textwidth]{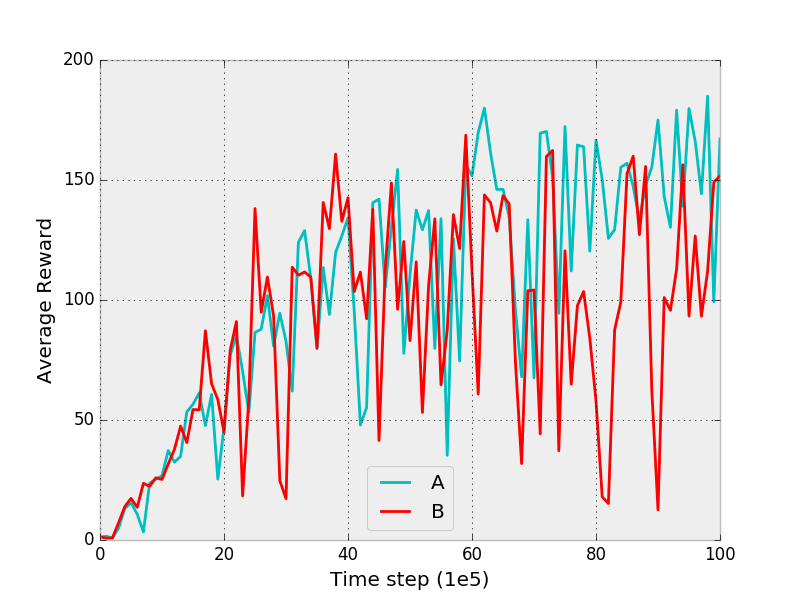}
    \caption{Determinism vs. replicability. A deterministic implementation executed on two separate machines produces different learning curves.}
    \label{detvrepl}
\end{figure*}

Figure \ref{detvrepl} demonstrates the distinction between deterministic implementations and replicability. It shows two learning curves of the same deterministic implementation executed on two different machines. Since the experimental conditions under which the deterministic implementation was executed is varied, we observe a discrepancy between the learning curves. The two machines had different GPU architectures as well as different operating systems and library versions, any or all of which may have elicited differences in the learning curves. Notice how the curves overlap in the very beginning as their network initializations are identical. When we execute the deterministic implementation on either machine repeatedly, it produces identical results. The results only vary when we compare across machines, or experimental conditions.

Note that the volatility in the learning curve is due to our use of a smaller DQN architecture \cite{dqn2013}. Our curves are consistent with the results we observe in that scenario \cite{dqn2013}.

\section{Comparison of Evaluation Protocols}
Our evaluation protocol in which the agent performs a greedy policy from 100 different start states ensures that any differences between the episode trajectories of two agents can be attributed to \textit{decisions made as consequences of their Q-networks being different}. This is a key feature we want in our evaluation protocol, because the differences in Q-networks can be attributed entirely to allowing an individual source of nondeterminism to influence the training process. In this way, we can safely claim that the variance in performance is solely caused by a source of nondeterminism. However, as we know, in the deterministic ALE ($p=0$), repeating a greedy policy 100 times results in 100 identical trajectories. Our evaluation protocol circumvents this issue and evaluates the agent in a diverse set of scenarios, through the use of 100 different start states.

Our evaluation protocol contrasts from the typical way of evaluating agents in diverse scenarios. Typically, diversity is produced by having agents perform $\epsilon$-greedy policies is during evaluations \cite{dqn2013,dqn}. However, in these evaluations, we are unable to attribute performance differences between the agents solely to differences between their Q-networks, since exploration can confound results. Even if exploration is seeded in the evaluation stage, a single deviation between policies will desynchronize the exploration seeds. If this occurs, then the differences in trajectories between two agents are influenced by the Q-networks \textit{and} exploratory actions. Our evaluation protocol allows us to correctly measure our variable of interest, i.e. sensitivity due to nondeterminism, while retaining the benefits of stochastic evaluation protocols.

\section{Generating Start Sequences}
We have two primary goals when generating start sequences. One goal is to ensure that our start states are diverse and are in different areas of the state space. In doing so, we can be confident that we are evaluating an agent's ability to generalize to different states. Another goal is to ensure that, in seeking diversity, we do not create poor start states that place an agent at a disadvantage. As stated in the main body of the paper, we generate predetermined action sequences for the agent to perform at the start of an episode, taking the agent to a unique state from which it performs a greedy policy for the remainder of the evaluation episode. The use of a predetermined action sequence is similar to using “human starts” at the beginning of episodes, where the agent completes an episode beginning with a trajectory of human expert play \cite{gorila}. However, rather than generating  our start sequences from human trajectories, we generate them computationally.  To do so, we first produce 1000 random action sequences, varying in length from 55 to 95 actions (chosen uniformly within this range). We find that for \textsc{Breakout}, choosing beyond 95 actions often results in very poor states, or the loss of a life. We wanted to maximize the lower bound on the number of random actions and informally found 55 to be a good number. The reason we vary the number of random actions is to improve the diversity of our start states. By varying the stage at which the agent performs greedy actions, it can help create more diverse states. However, by using between 55 and 95 random actions, we may find several states to be poor, as we might expect from a long sequence of random actions. To remedy this, we used a trained DQN to rank our 1000 generated start states by their maximum Q-value (i.e. $\max_{a}Q(s,a)$). We then select 100 start sequences randomly from the top 250 start sequences. While this method of generating start sequences is biased towards generating sequences that a DQN may perform well on, it enables us to generate longer sequences of random actions, which improves the diversity of our start states. We use the same 100 start sequences at every evaluation interval for all agents (except for the environment group). We should note that, while no-op actions are often used at the beginning of an episode, we did not explicitly use them in our start state generation procedure since no-op actions have no effect at the beginning of \textsc{Breakout} episodes. This is a known drawback to no-ops, and \textsc{Breakout} is not the only domain for which no-ops have no effect at the beginning of the episode \cite{revisitingale}. We include our start sequences along with the provided code linked in the main body of the paper.

\section{Experimental Conditions}
We list here the experimental conditions used for our experiments. The hardware conditions were:
\begin{itemize}
    \item GPU: Nvidia GeForce GTX 1080
    \item CPUs (12)
    \begin{itemize}
        \item Model: Intel(R) Xeon(R) CPU E5-2603 v4 \@ 1.70GHz
        \item Architecture: Intel x86\_64 
        \item CPU op-mode(s): 32-bit, 64-bit
    \end{itemize}
\end{itemize}

Our software versions were: 
\begin{itemize}
    \item Python 2.7
    \item ALE (0.5.1)
    \item numpy (1.13.3)
    \item torch (0.3.0.post4) (PyTorch)
    \item torchvision (0.2.0)
    \item Operating system: Ubuntu 16.04 (xenial)
    \item Cuda (8.0.61)
    \item cuDNN (7003)
    \item Nvidia GPU Driver Version: 384.111
\end{itemize}

It is difficult to determine with absolute certainty the list of all experimental conditions that must necessarily be fixed to achieve replicability. One reason for this difficulty is that the experimental conditions that must be fixed are dependent on the implementation. Furthermore, to determine whether a specific experimental condition is necessary for replicability, we must be able to modify that experimental condition while fixing all others to see if it produces different results. This is an intractable approach, unfortunately. For example, suppose the specific version of the ALE used is irrelevant for achieving replicability. In order to conclude with certainty that the ALE version is not an experimental condition that must be fixed, we have to verify that for \textit{all} ALE versions the results are not changed. We must do this for all software that we use and we may encounter dependency issues or incompatibility. It is even more difficult to test which hardware conditions are necessary for replicability, because one cannot easily hold all hardware conditions fixed and switch the GPU that is used by a machine, for example. Furthermore, there are many hardware conditions that would need to be tested, most of which are irrelevant.

Despite the difficulty in determining the necessary conditions for replicability, we are reasonably confident that if the experimental conditions listed above are fixed, then our results will be replicated. In this paper, we sought to investigate the benefits of deterministic implementations, one of which is replicability. However, replicability was not our primary goal. If one wishes to develop a replicable experiment, we recommend that one develop a deterministic implementation (as we do in this paper) and combine it with the tools listed in the related work section.

\section{Start State Evaluations}

In Figure \ref{indiecurvescomplete} we include the learning curves of a deterministic implementation of deep Q-learning. Each graph represents the agent's score after performing a specific action sequence to start the episode. The key feature consistently observable across all of these curves is the volatility of the agent's performance. We see that as the agent learns, its performance on individual start states fluctuates heavily. The general trend of increasing performance we see for a DQN is based off of an average of these 100 curves. However, even if we observe a general improvement in average score, the agent becomes drastically worse on specific start sequences. Standard DRL evaluations are stochastic, causing different evaluations at each evaluation interval. Given the volatility observed in Figure \ref{indiecurvescomplete}, perhaps it is best to use deterministic evaluations. Allowing evaluations to vary across evaluation intervals might not accurately measure the agent's change in performance, a problem that can be mitigated with deterministic evaluations.

\captionsetup[subfigure]{subrefformat=simple,labelformat=simple,listofformat=subsimple}
\renewcommand\thesubfigure{(\alphalph{\value{subfigure}})}
\begin{figure}[ht]
\centering
	\subfloat[][]{
    	\includegraphics[width=0.25\linewidth]{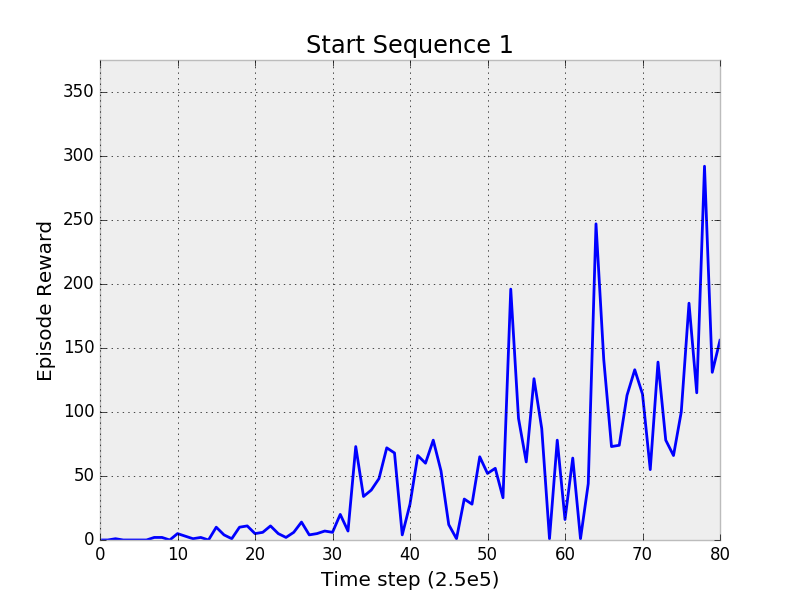}}
    \subfloat[][]{
    	\includegraphics[width=0.25\linewidth]{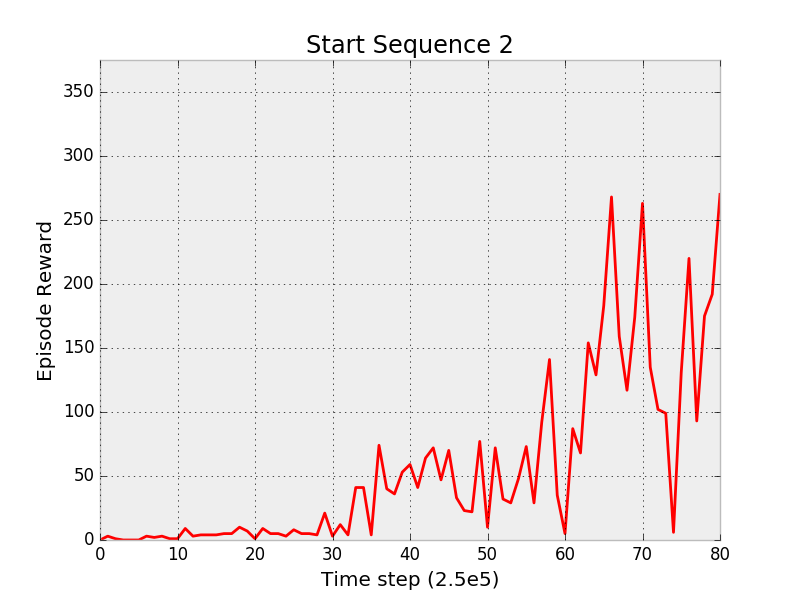}}
	\subfloat[][]{
    	\includegraphics[width=0.25\linewidth]{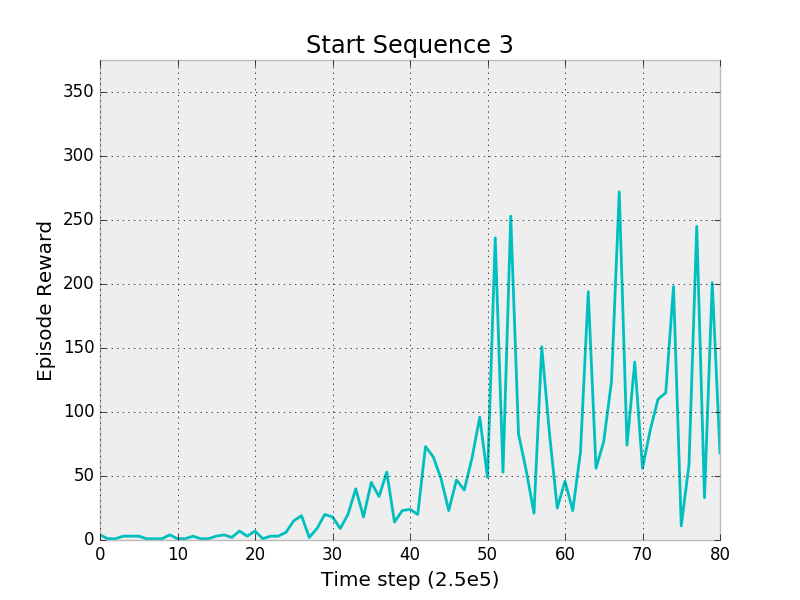}}
    \subfloat[][]{
    	\includegraphics[width=0.25\linewidth]{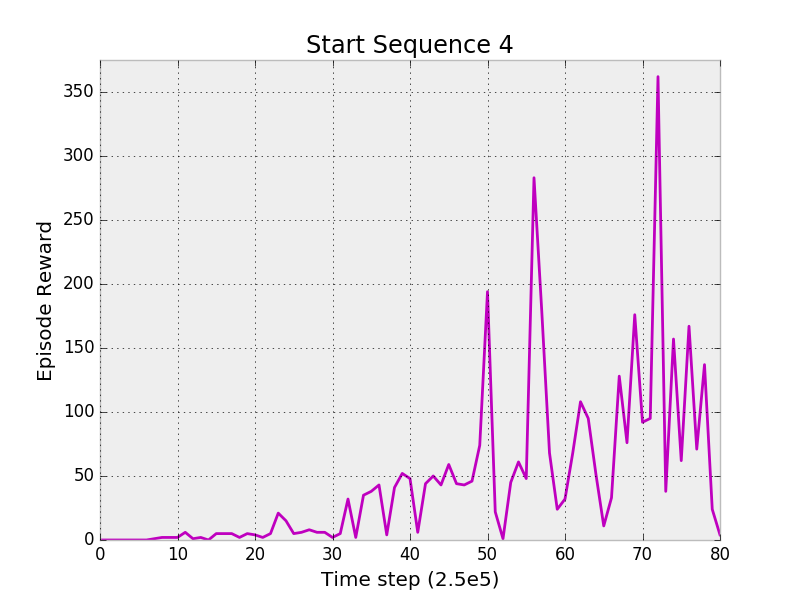}} \\
	\subfloat[][]{
    	\includegraphics[width=0.25\linewidth]{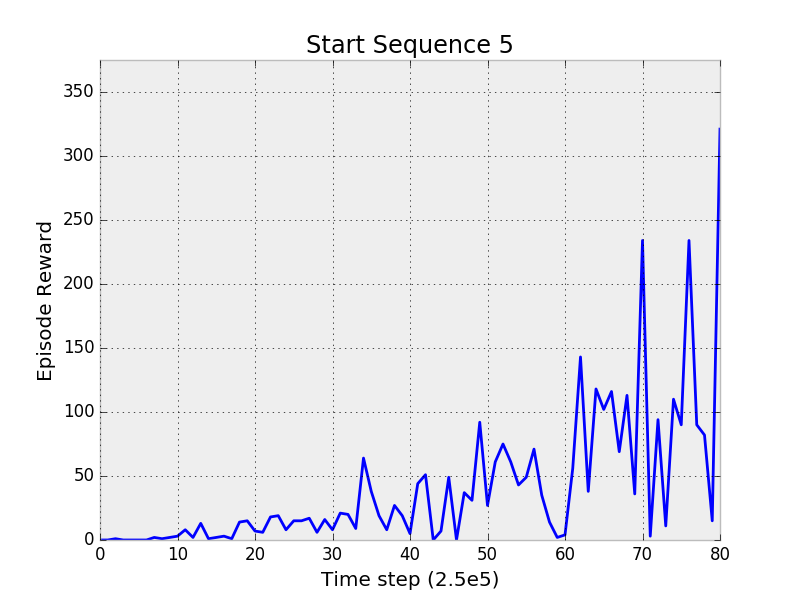}}
    \subfloat[][]{
    	\includegraphics[width=0.25\linewidth]{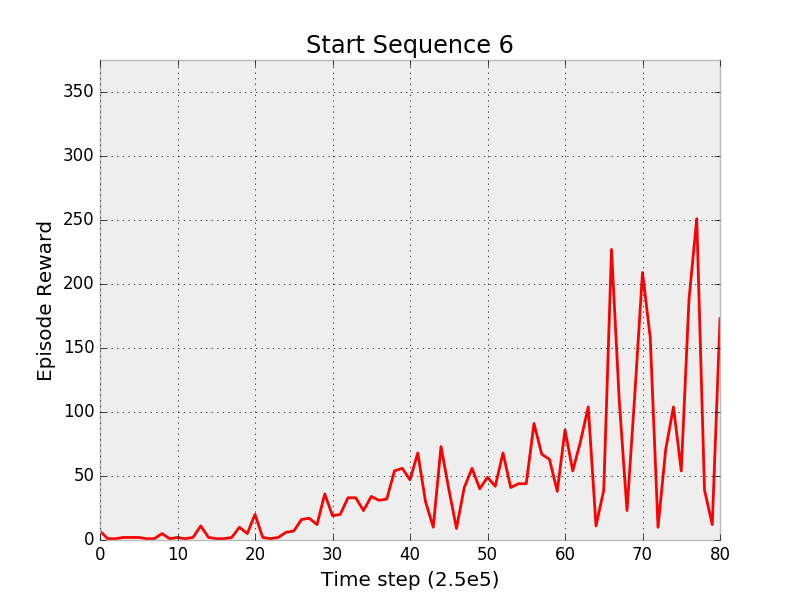}}
	\subfloat[][]{
    	\includegraphics[width=0.25\linewidth]{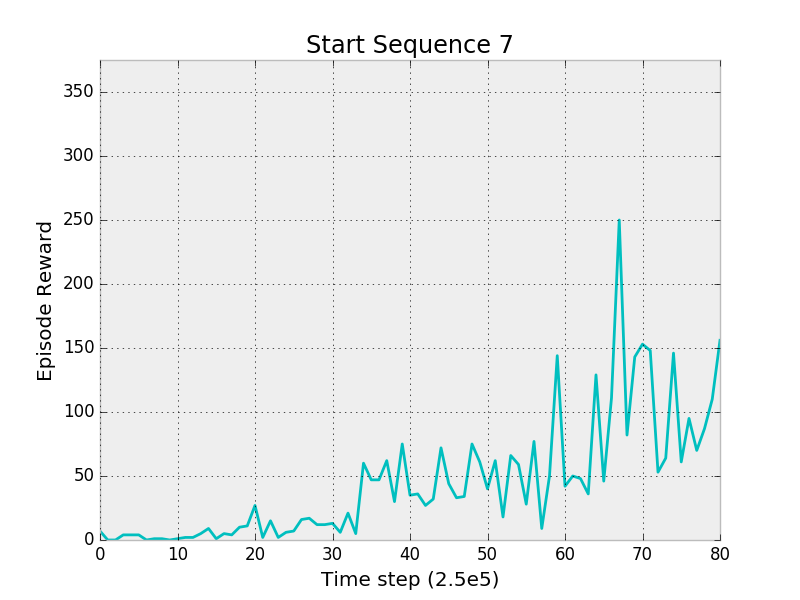}}
    \subfloat[][]{
    	\includegraphics[width=0.25\linewidth]{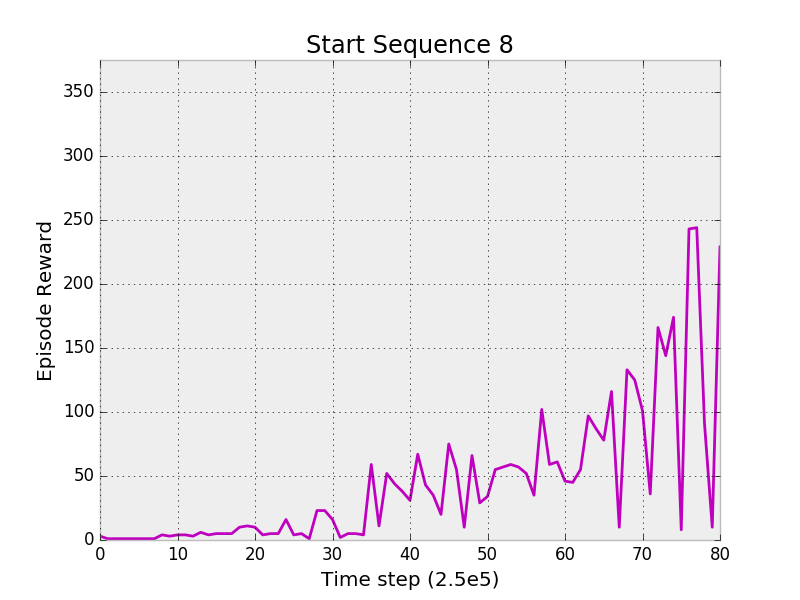}} \\
	\subfloat[][]{
    	\includegraphics[width=0.25\linewidth]{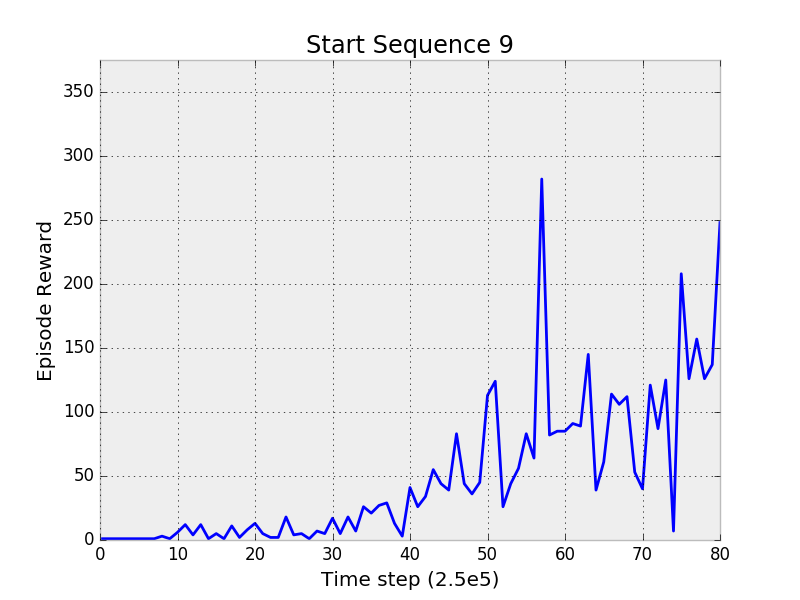}}
    \subfloat[][]{
    	\includegraphics[width=0.25\linewidth]{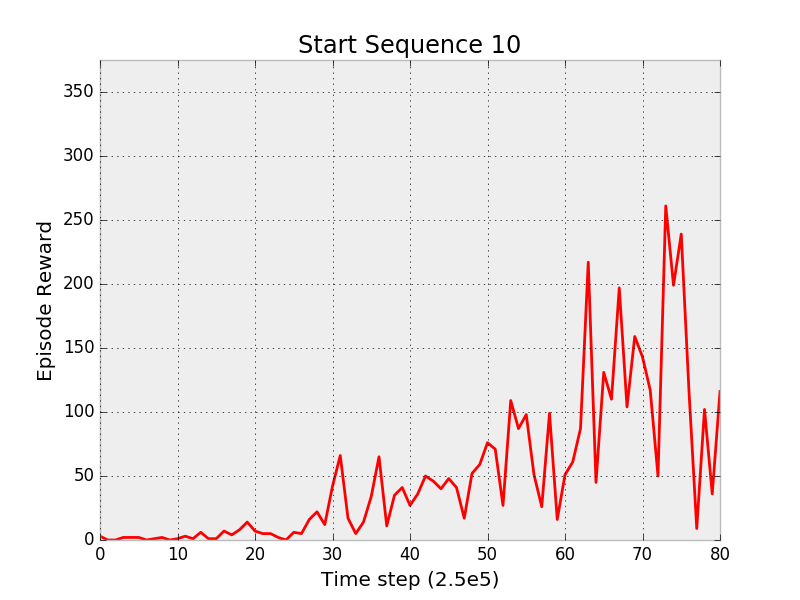}}
	\subfloat[][]{
    	\includegraphics[width=0.25\linewidth]{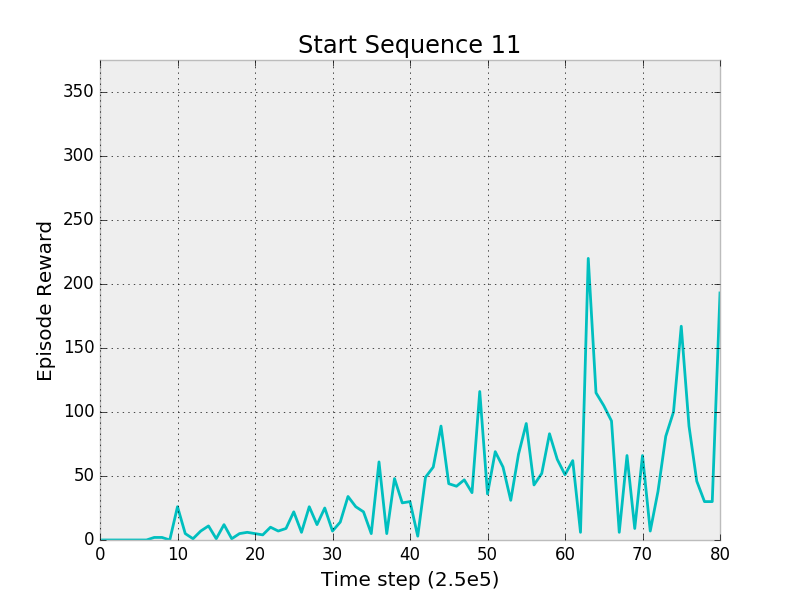}}
    \subfloat[][]{
    	\includegraphics[width=0.25\linewidth]{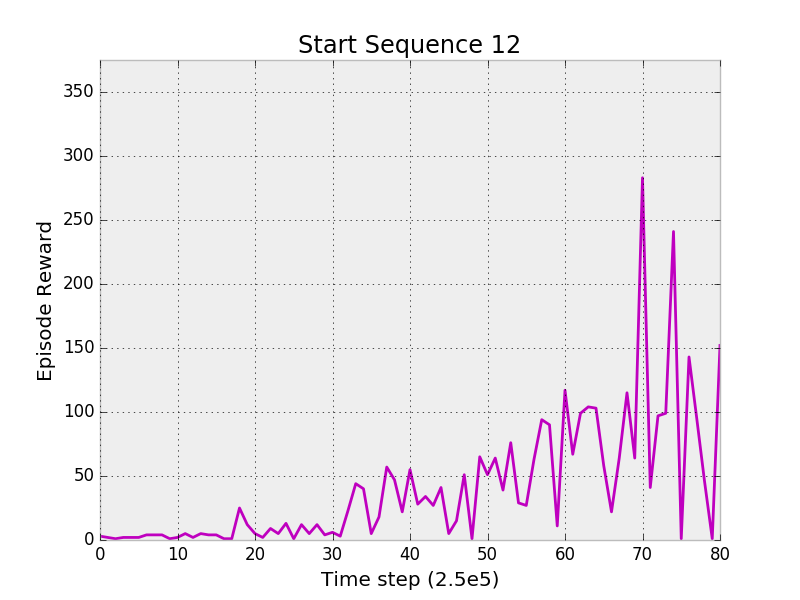}} \\
	\subfloat[][]{
    	\includegraphics[width=0.25\linewidth]{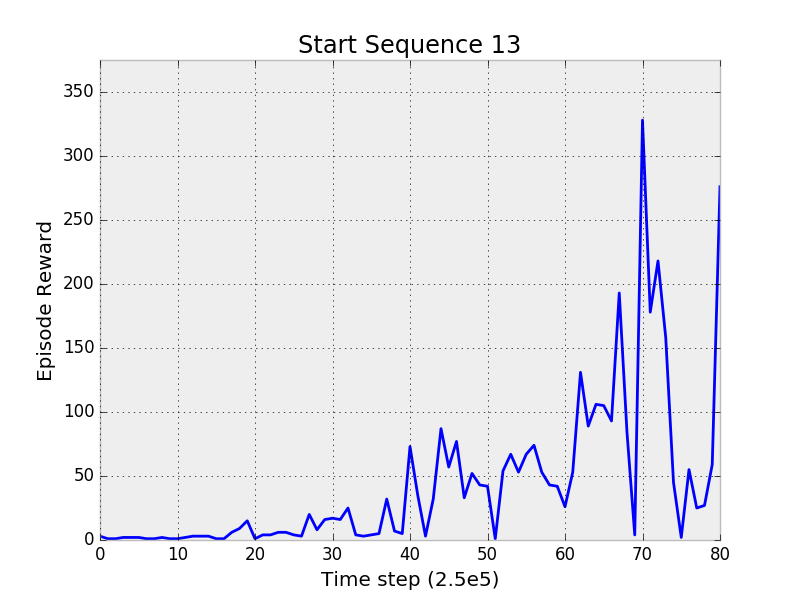}}
    \subfloat[][]{
    	\includegraphics[width=0.25\linewidth]{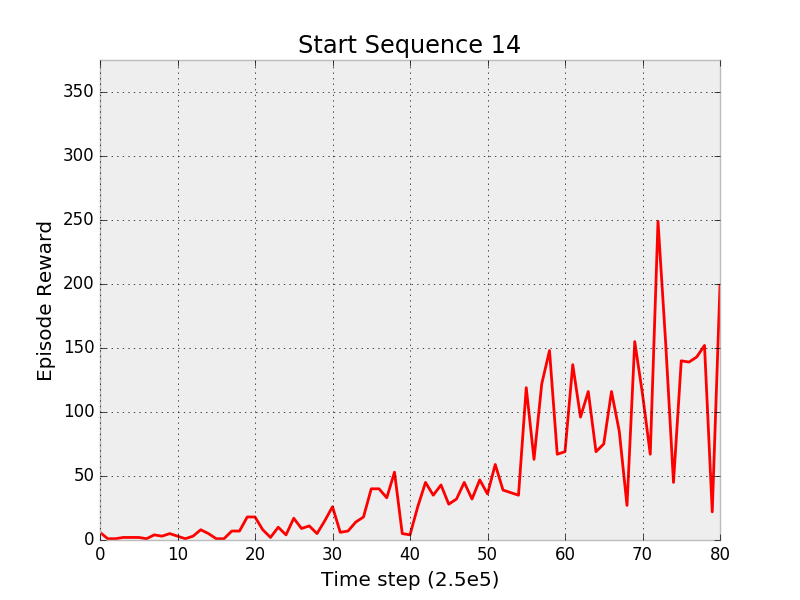}}
	\subfloat[][]{
    	\includegraphics[width=0.25\linewidth]{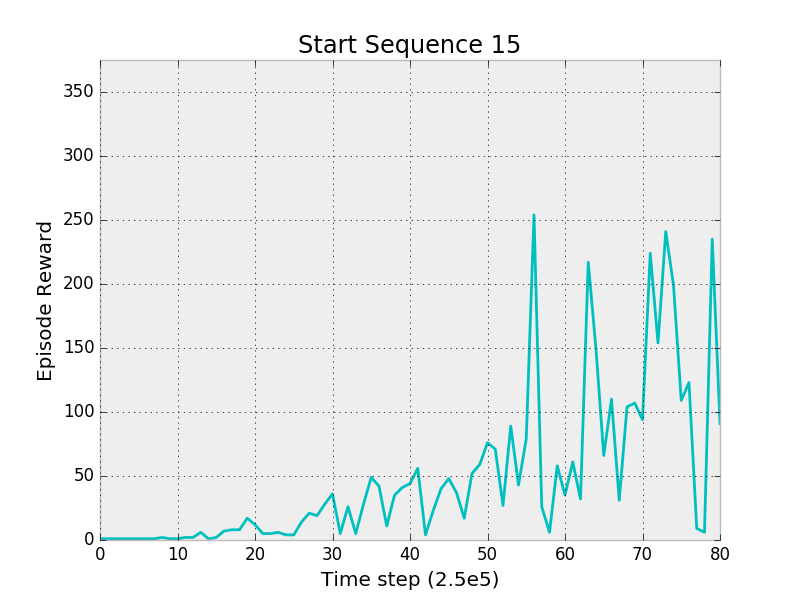}}
    \subfloat[][]{
    	\includegraphics[width=0.25\linewidth]{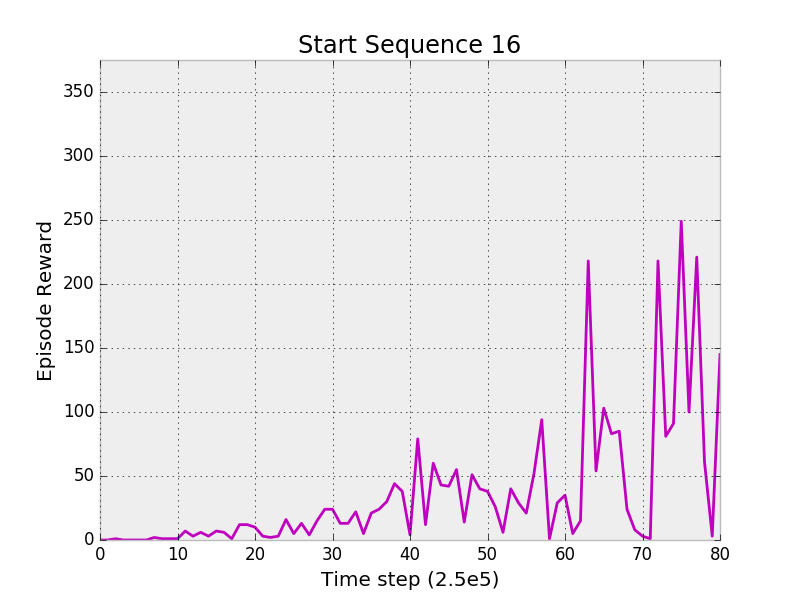}} \\
	\subfloat[][]{
    	\includegraphics[width=0.25\linewidth]{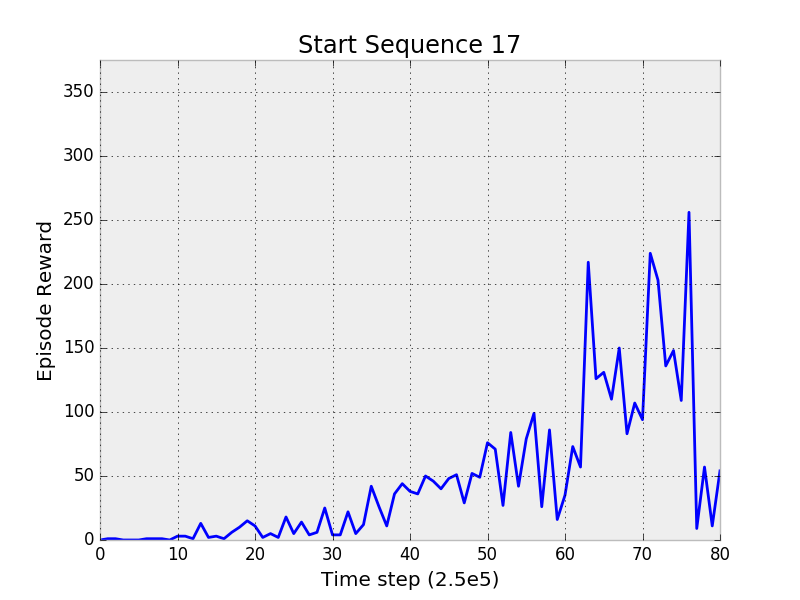}}
    \subfloat[][]{
    	\includegraphics[width=0.25\linewidth]{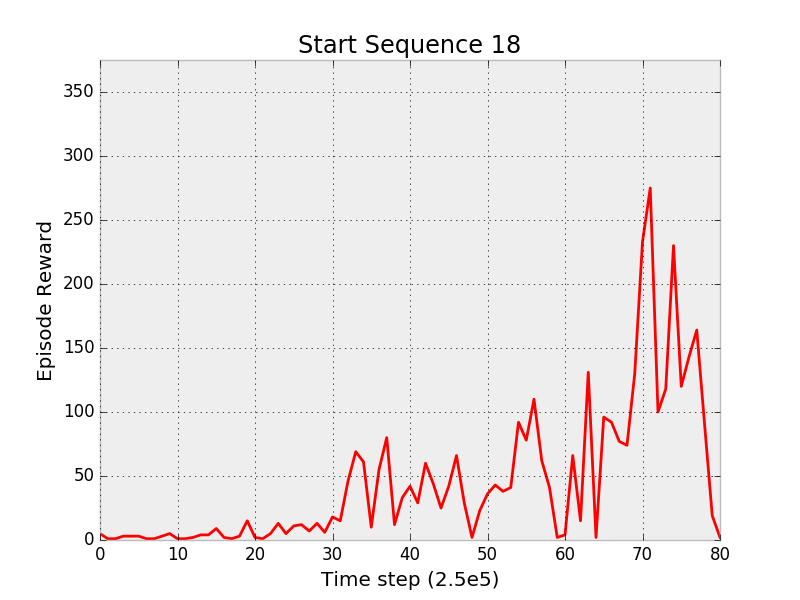}}
	\subfloat[][]{
    	\includegraphics[width=0.25\linewidth]{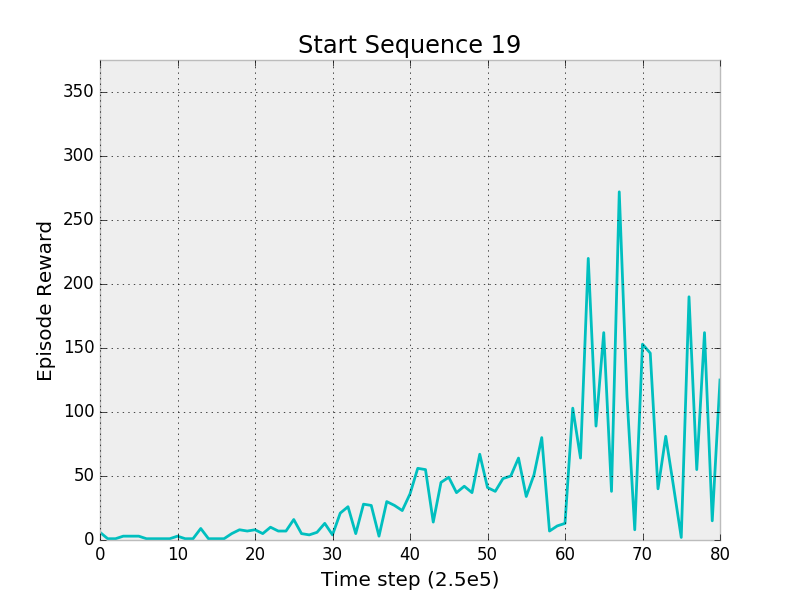}}
    \subfloat[][]{
    	\includegraphics[width=0.25\linewidth]{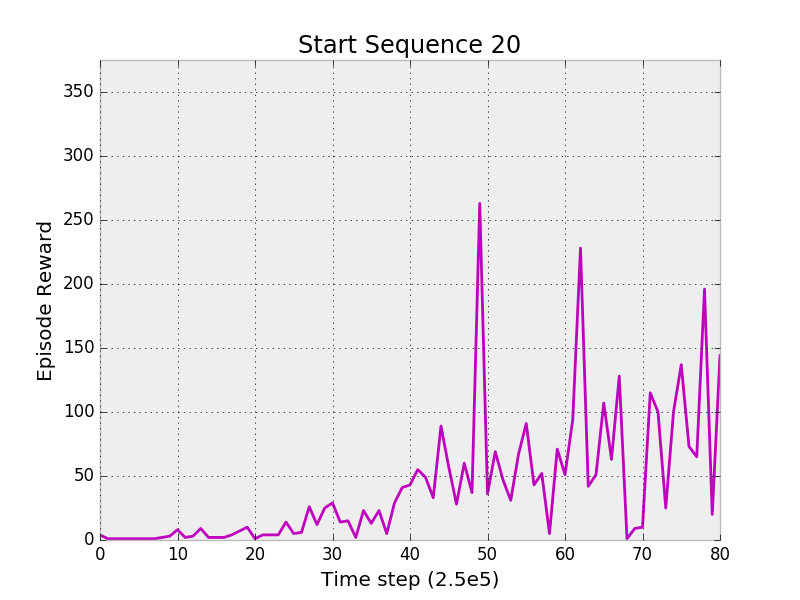}} \\
\caption{Learning curves of the deterministic agent for individual start states/sequences.}
\label{indiecurvescomplete}
\end{figure}

\begin{figure}[ht]
\ContinuedFloat
\centering
	\subfloat[][]{
    	\includegraphics[width=0.25\linewidth]{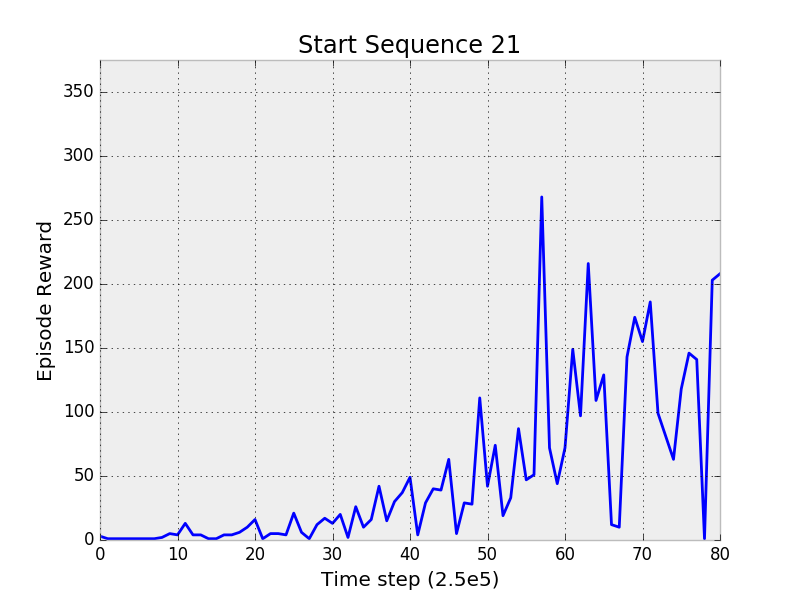}}
    \subfloat[][]{
    	\includegraphics[width=0.25\linewidth]{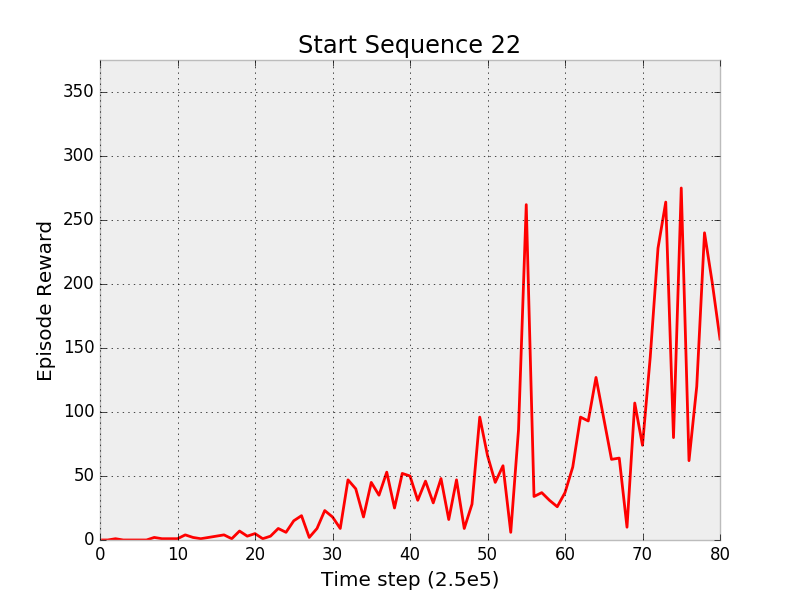}}
	\subfloat[][]{
    	\includegraphics[width=0.25\linewidth]{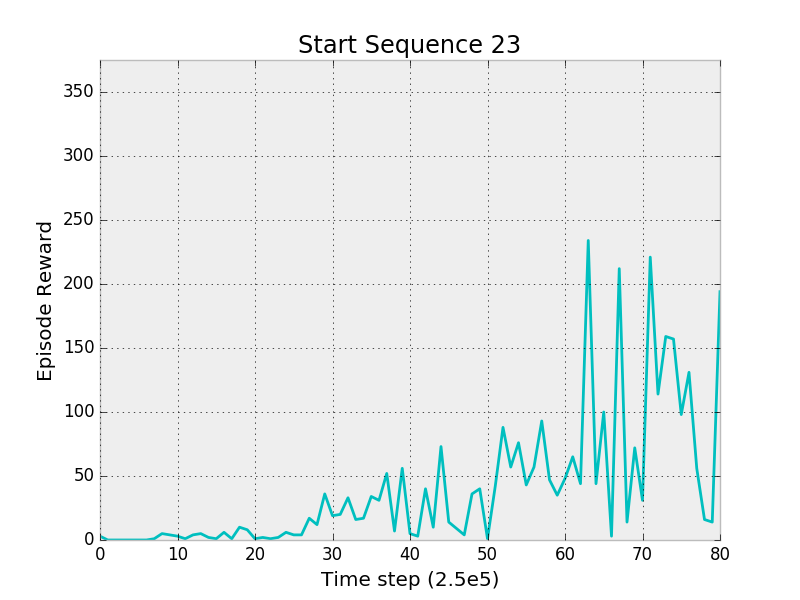}}
    \subfloat[][]{
    	\includegraphics[width=0.25\linewidth]{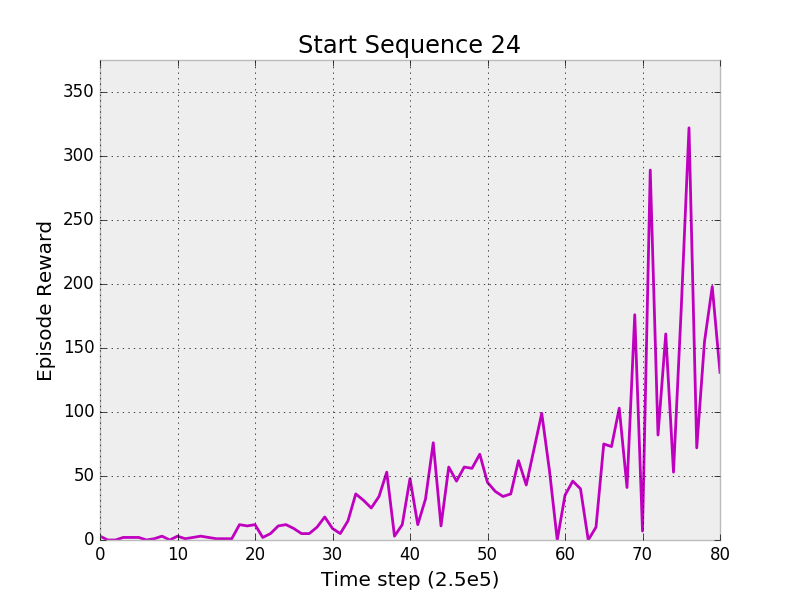}} \\
	\subfloat[][]{
    	\includegraphics[width=0.25\linewidth]{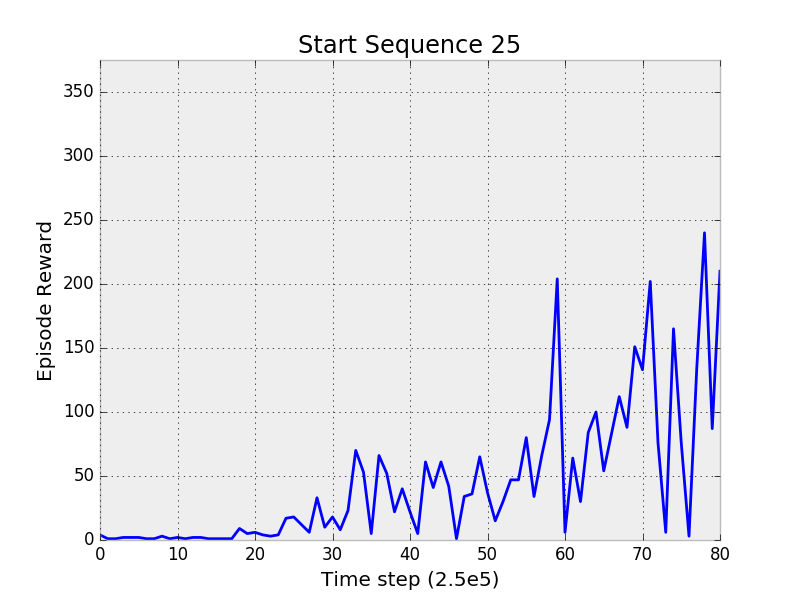}}
    \subfloat[][]{
    	\includegraphics[width=0.25\linewidth]{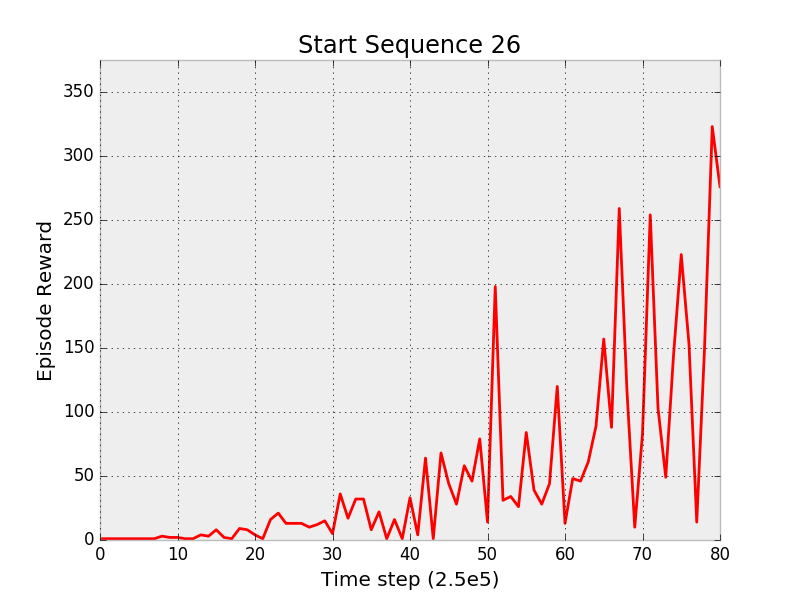}}
	\subfloat[][]{
    	\includegraphics[width=0.25\linewidth]{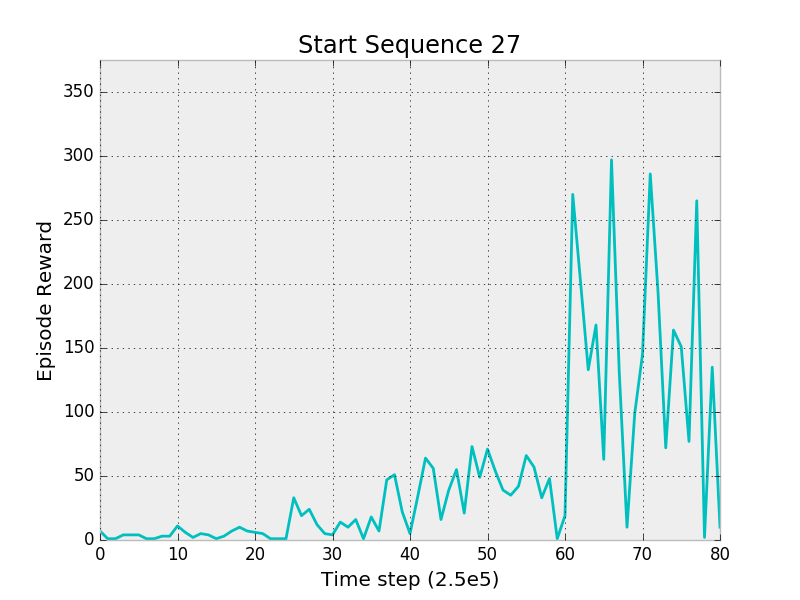}}
    \subfloat[][]{
    	\includegraphics[width=0.25\linewidth]{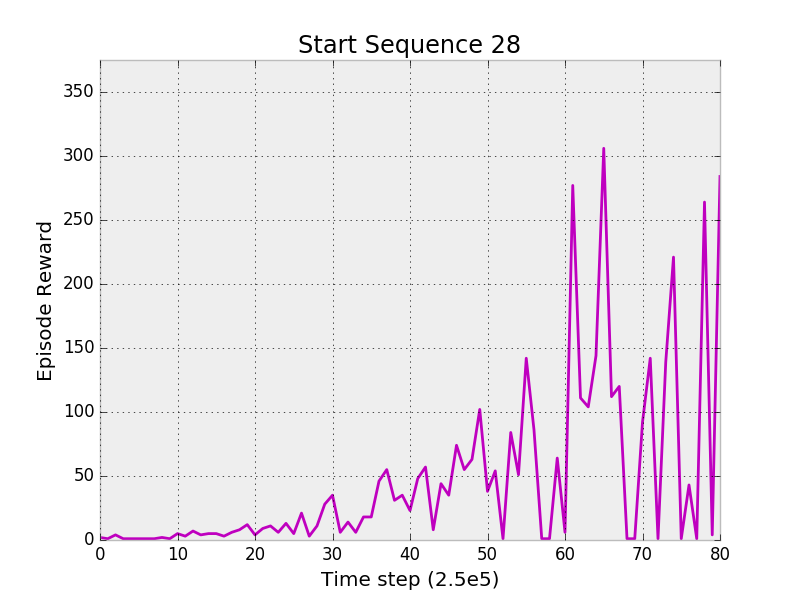}} \\
	\subfloat[][]{
    	\includegraphics[width=0.25\linewidth]{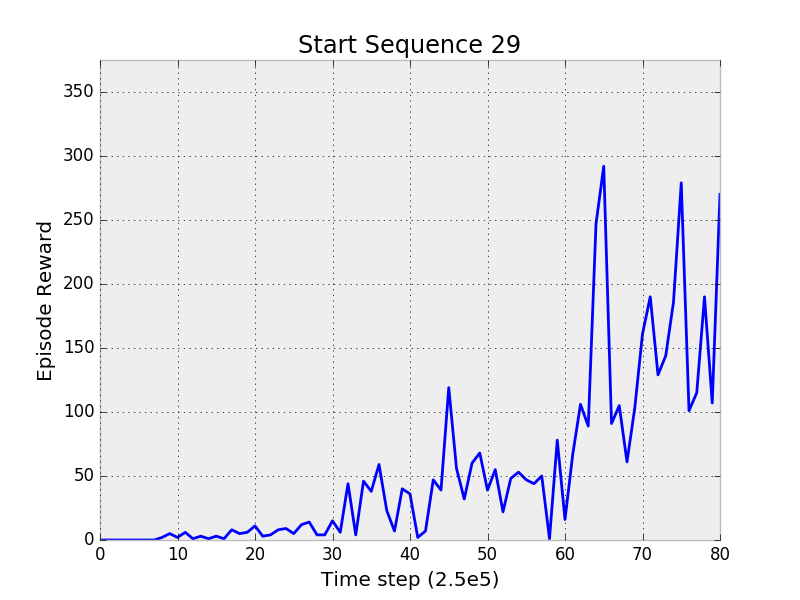}}
    \subfloat[][]{
    	\includegraphics[width=0.25\linewidth]{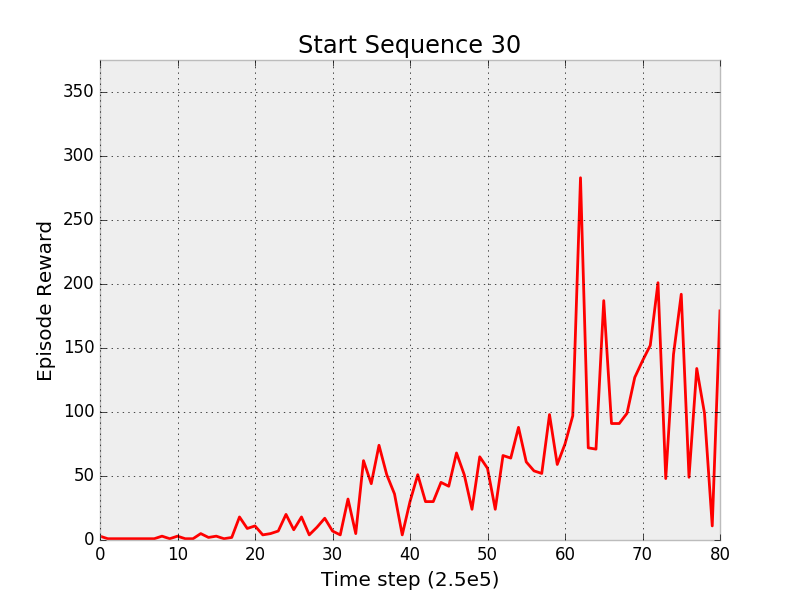}}
	\subfloat[][]{
    	\includegraphics[width=0.25\linewidth]{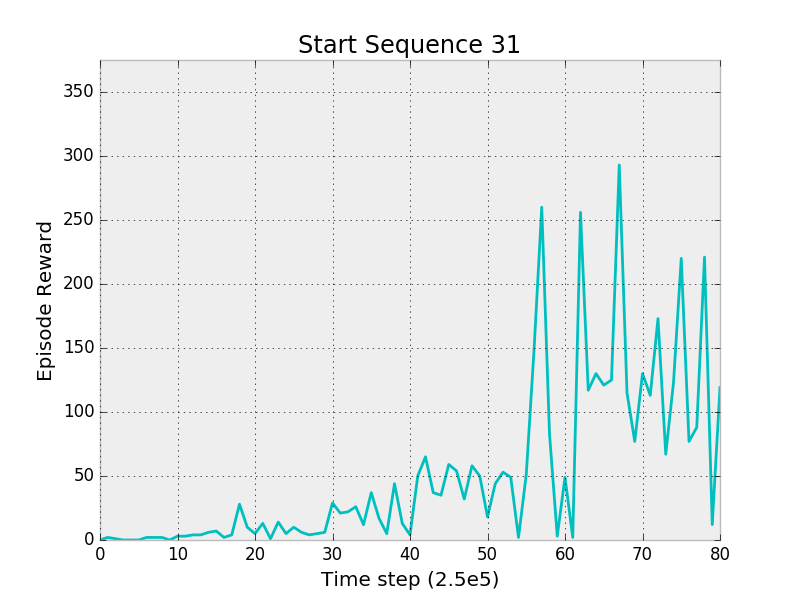}}
    \subfloat[][]{
    	\includegraphics[width=0.25\linewidth]{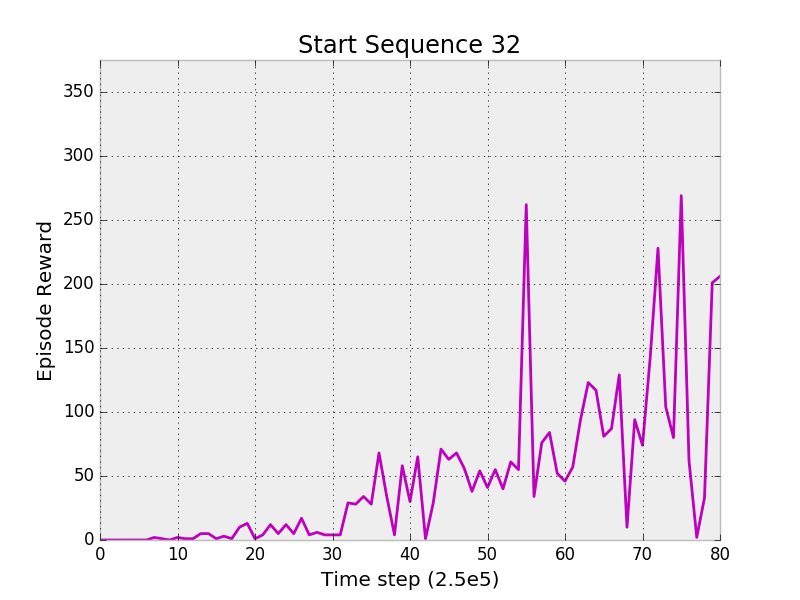}} \\
	\subfloat[][]{
    	\includegraphics[width=0.25\linewidth]{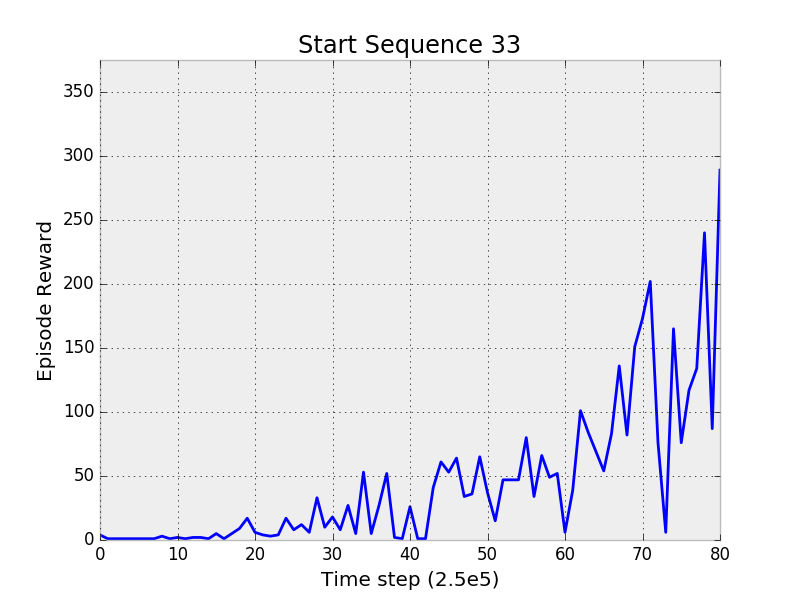}}
    \subfloat[][]{
    	\includegraphics[width=0.25\linewidth]{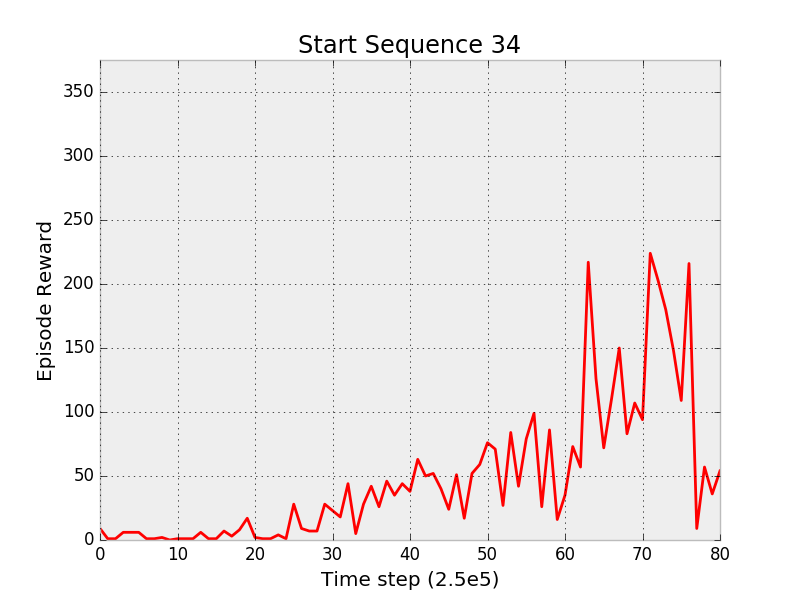}}
	\subfloat[][]{
    	\includegraphics[width=0.25\linewidth]{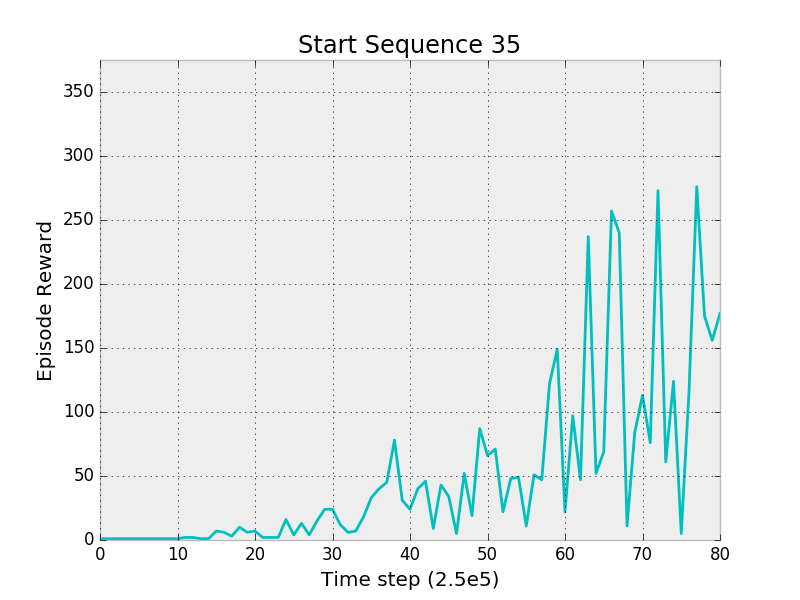}}
    \subfloat[][]{
    	\includegraphics[width=0.25\linewidth]{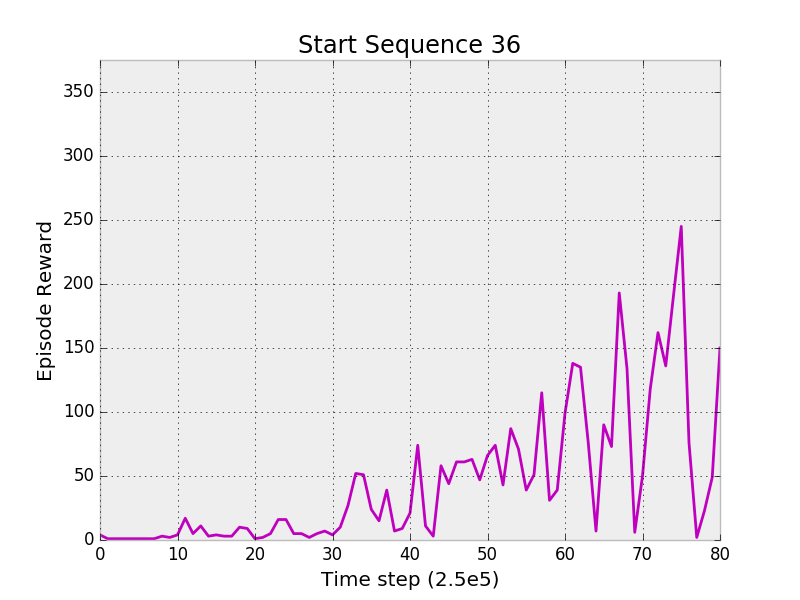}} \\
	\subfloat[][]{
    	\includegraphics[width=0.25\linewidth]{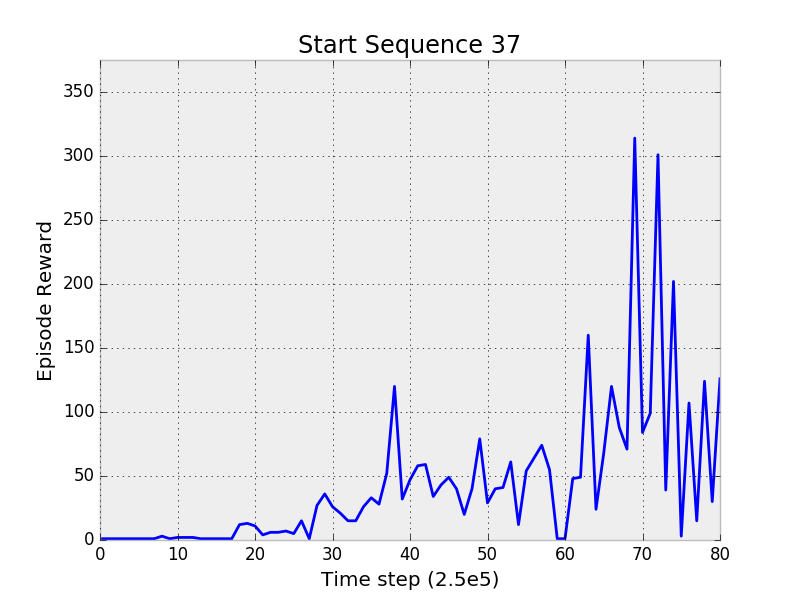}}
    \subfloat[][]{
    	\includegraphics[width=0.25\linewidth]{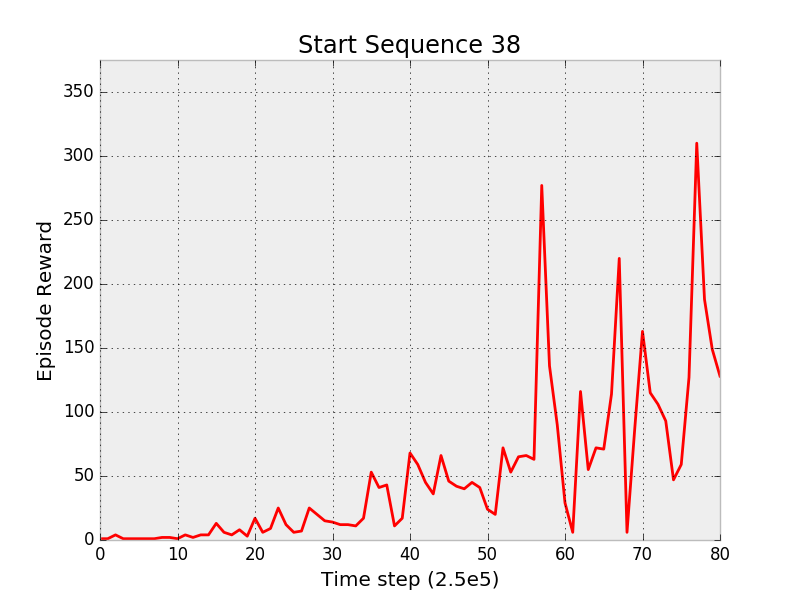}}
	\subfloat[][]{
    	\includegraphics[width=0.25\linewidth]{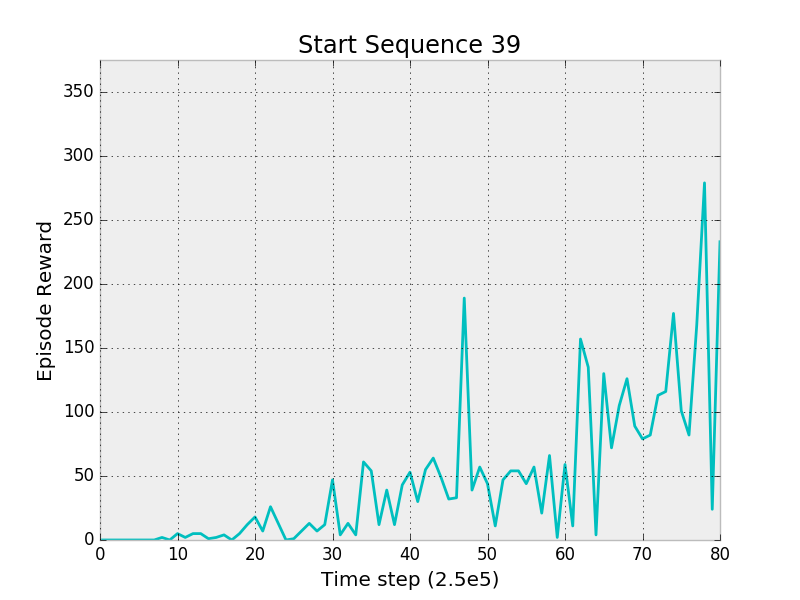}}
    \subfloat[][]{
    	\includegraphics[width=0.25\linewidth]{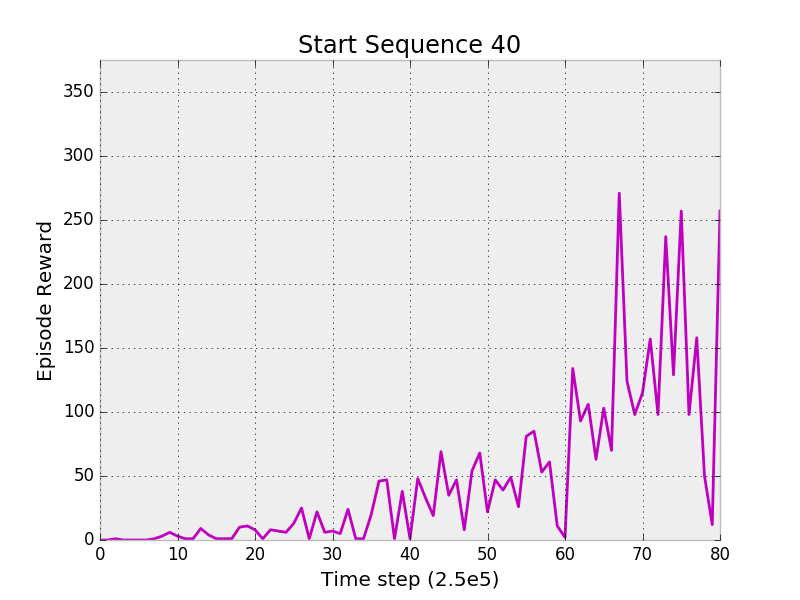}} \\
\caption{Learning curves of the deterministic agent for individual start states/sequences.}
\end{figure}

\begin{figure}[ht]
\ContinuedFloat
\centering
	\subfloat[][]{
    	\includegraphics[width=0.25\linewidth]{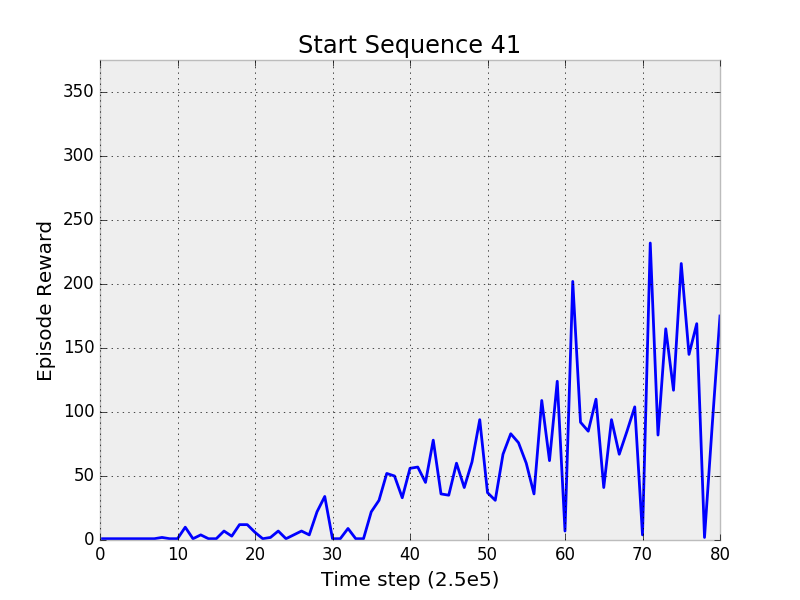}}
    \subfloat[][]{
    	\includegraphics[width=0.25\linewidth]{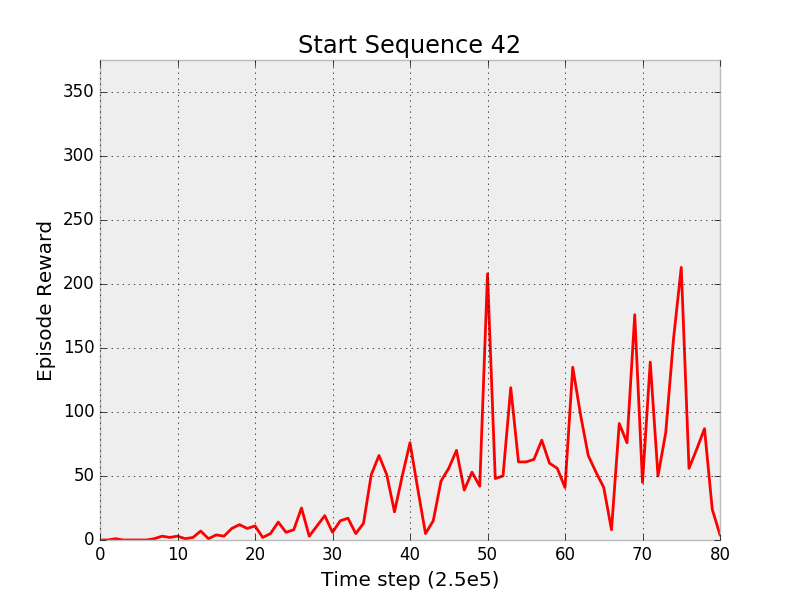}}
	\subfloat[][]{
    	\includegraphics[width=0.25\linewidth]{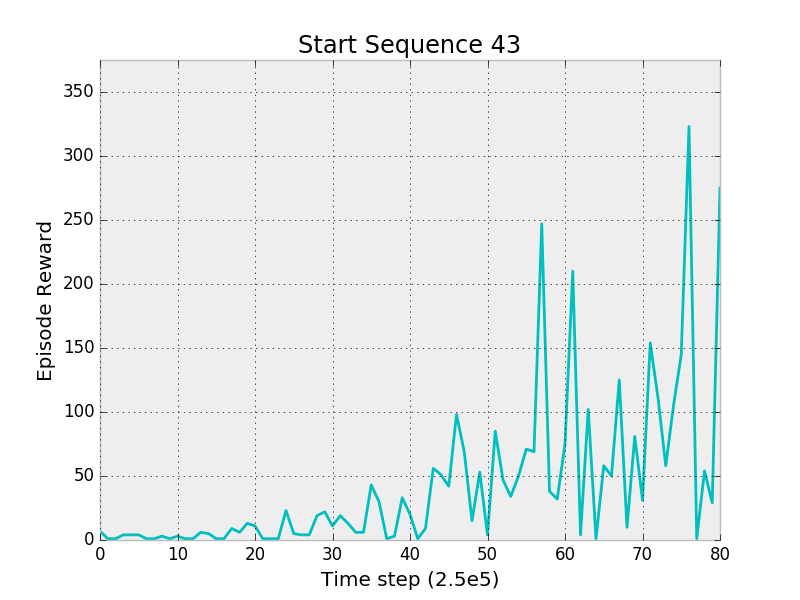}}
    \subfloat[][]{
    	\includegraphics[width=0.25\linewidth]{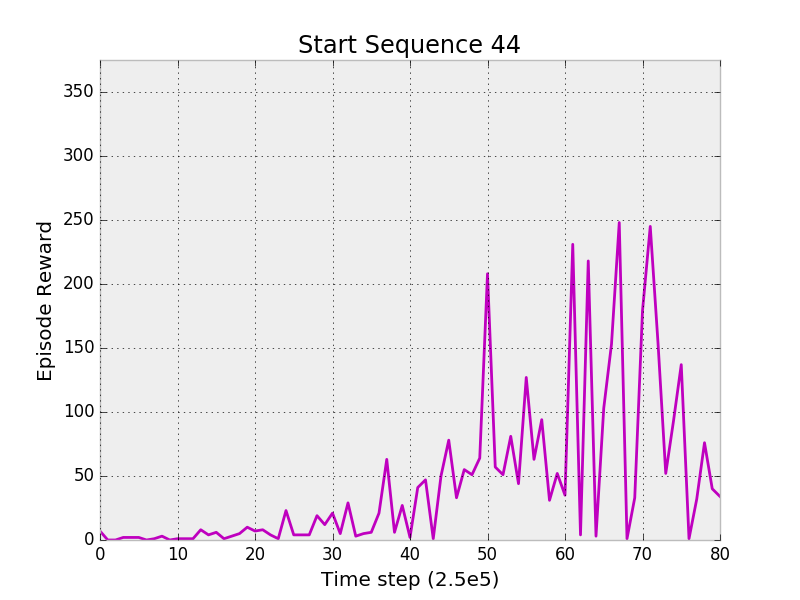}} \\
	\subfloat[][]{
    	\includegraphics[width=0.25\linewidth]{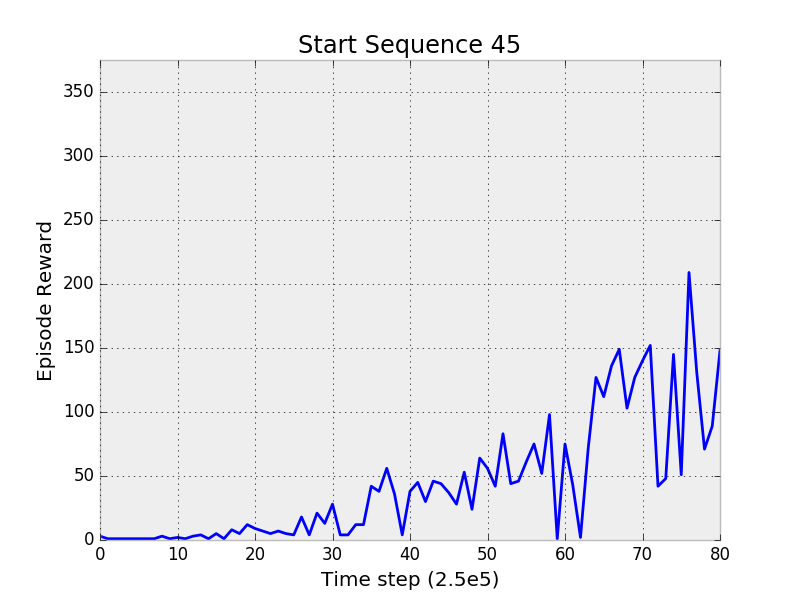}}
    \subfloat[][]{
    	\includegraphics[width=0.25\linewidth]{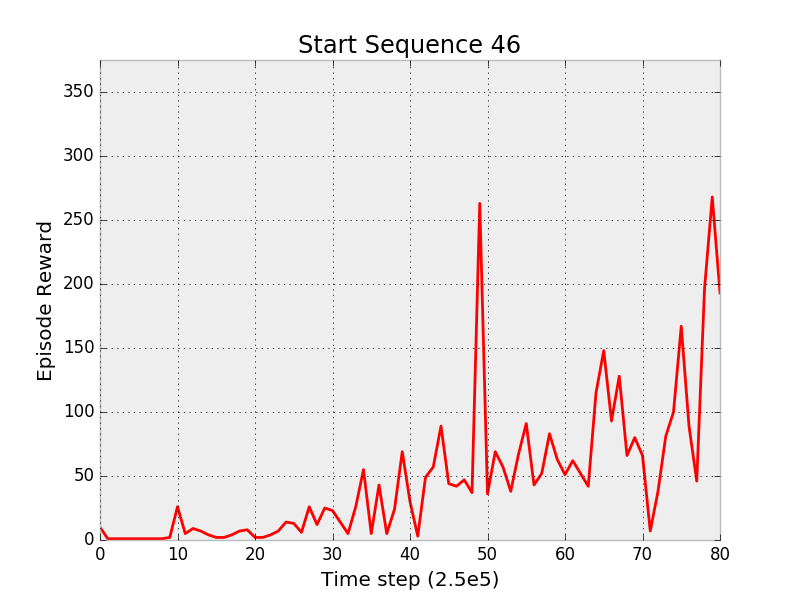}}
	\subfloat[][]{
    	\includegraphics[width=0.25\linewidth]{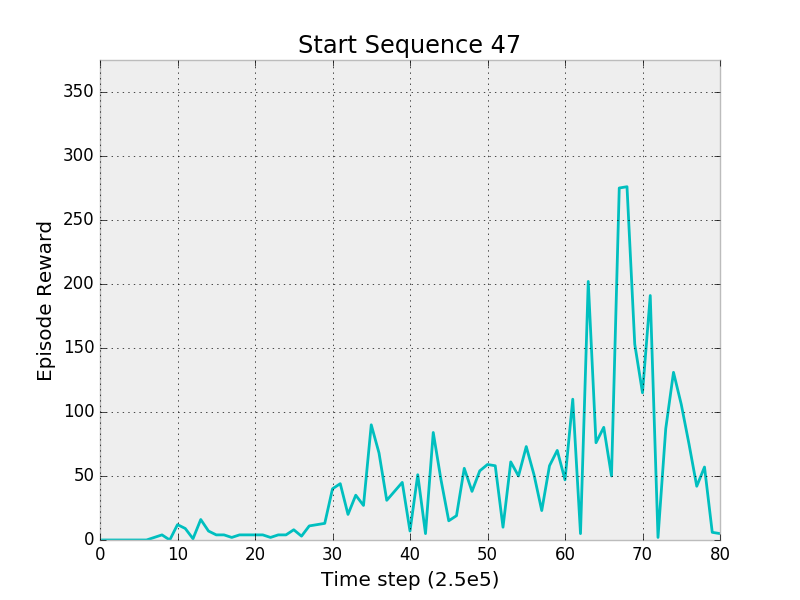}}
    \subfloat[][]{
    	\includegraphics[width=0.25\linewidth]{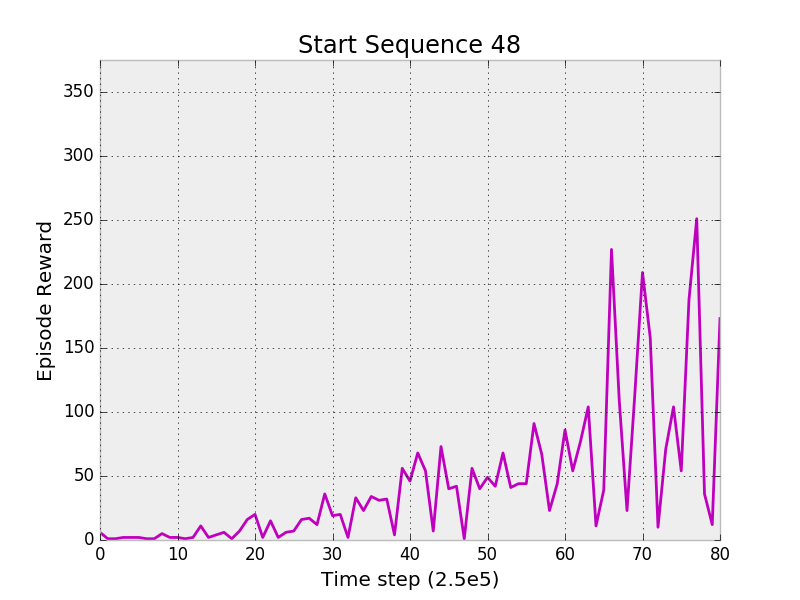}} \\
	\subfloat[][]{
    	\includegraphics[width=0.25\linewidth]{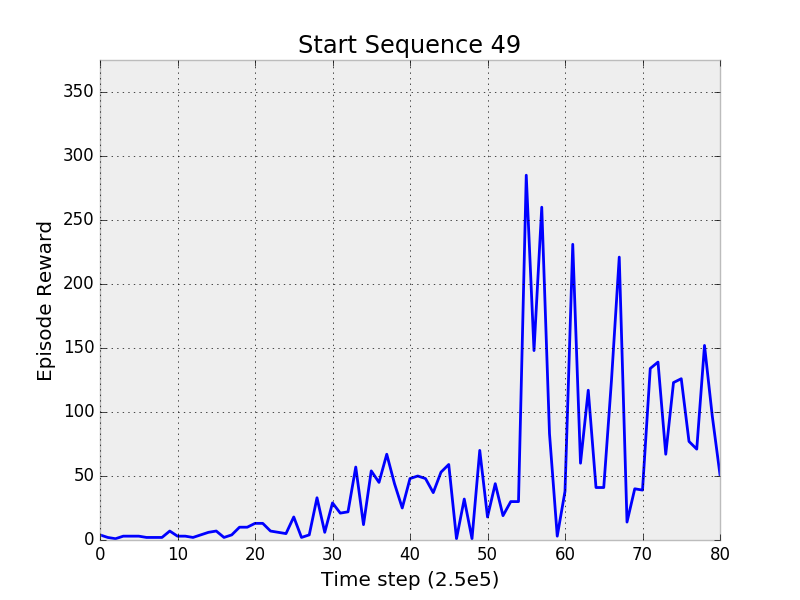}}
    \subfloat[][]{
    	\includegraphics[width=0.25\linewidth]{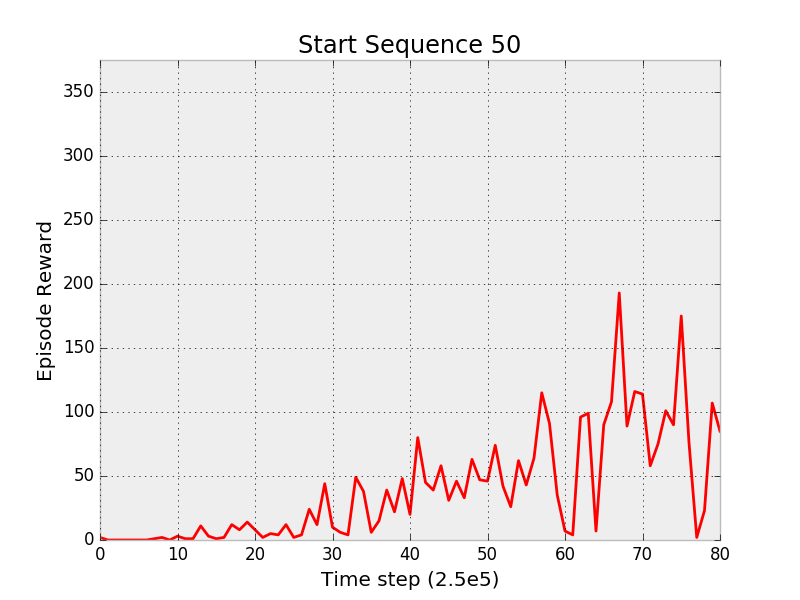}}
	\subfloat[][]{
    	\includegraphics[width=0.25\linewidth]{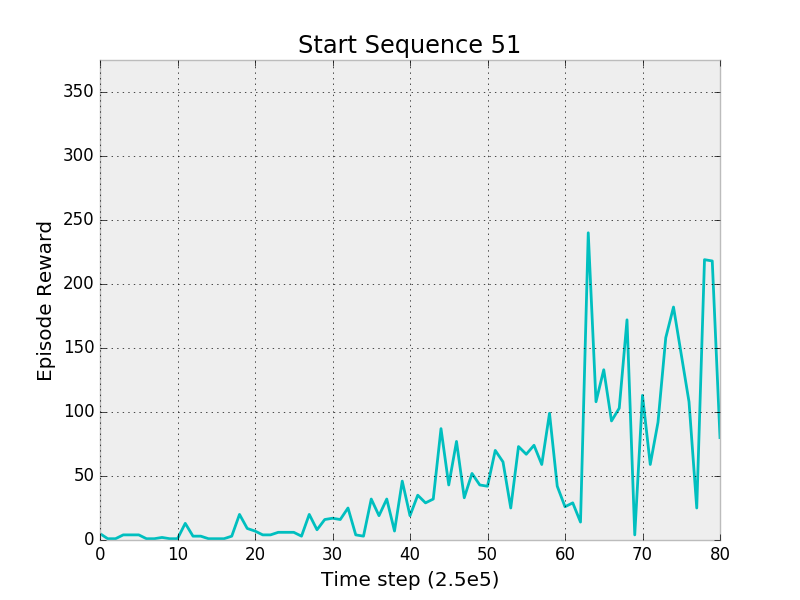}}
    \subfloat[][]{
    	\includegraphics[width=0.25\linewidth]{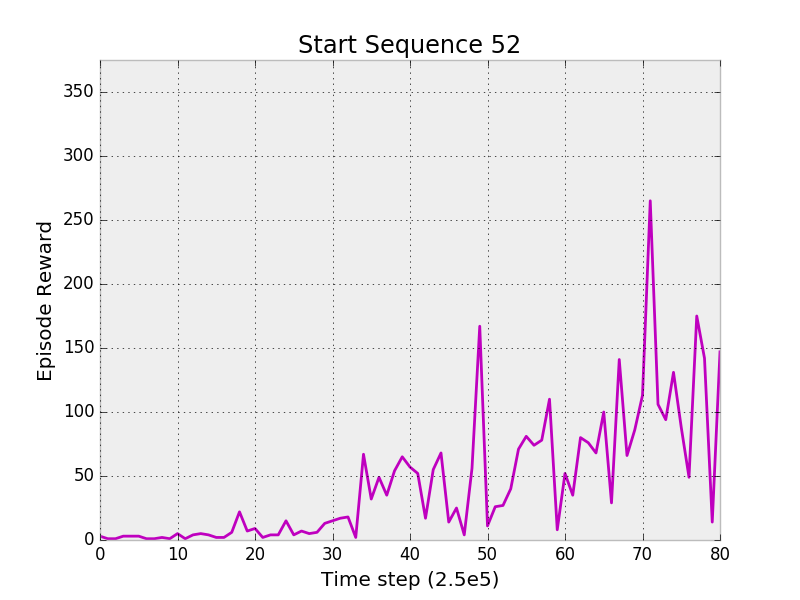}} \\
	\subfloat[][]{
    	\includegraphics[width=0.25\linewidth]{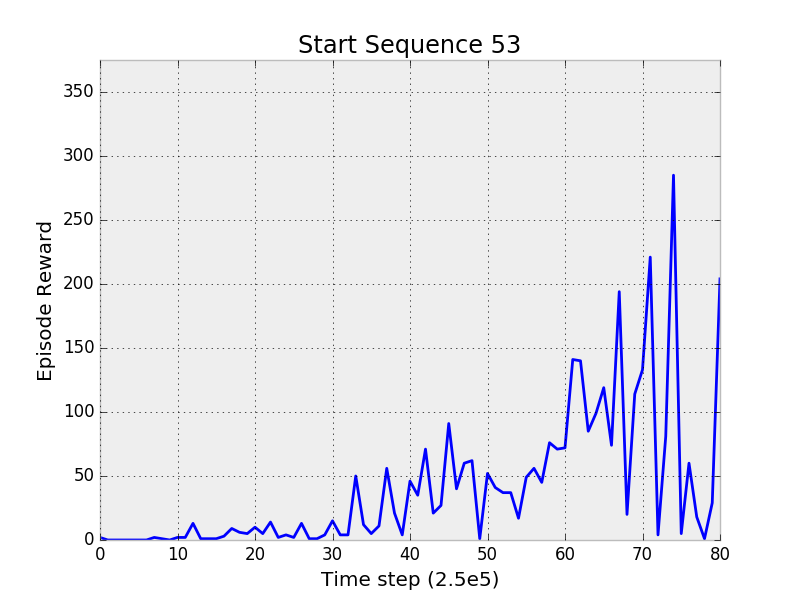}}
    \subfloat[][]{
    	\includegraphics[width=0.25\linewidth]{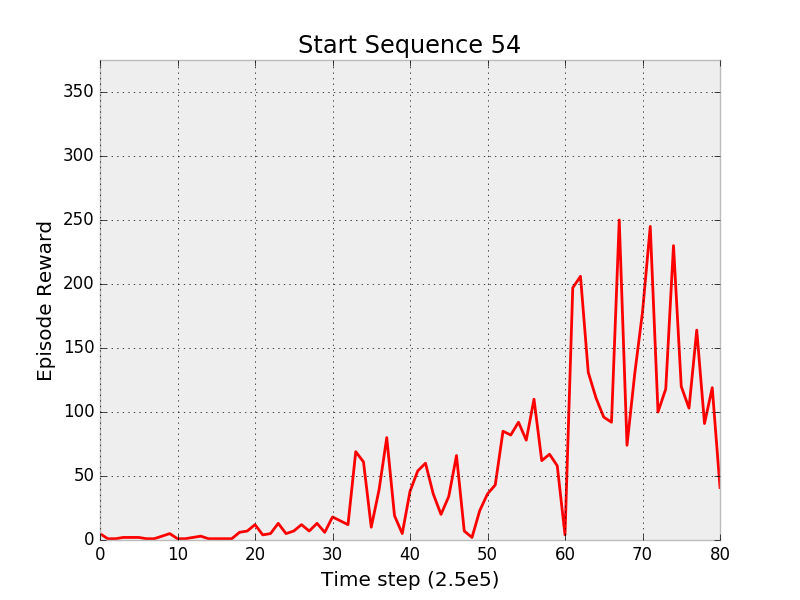}}
	\subfloat[][]{
    	\includegraphics[width=0.25\linewidth]{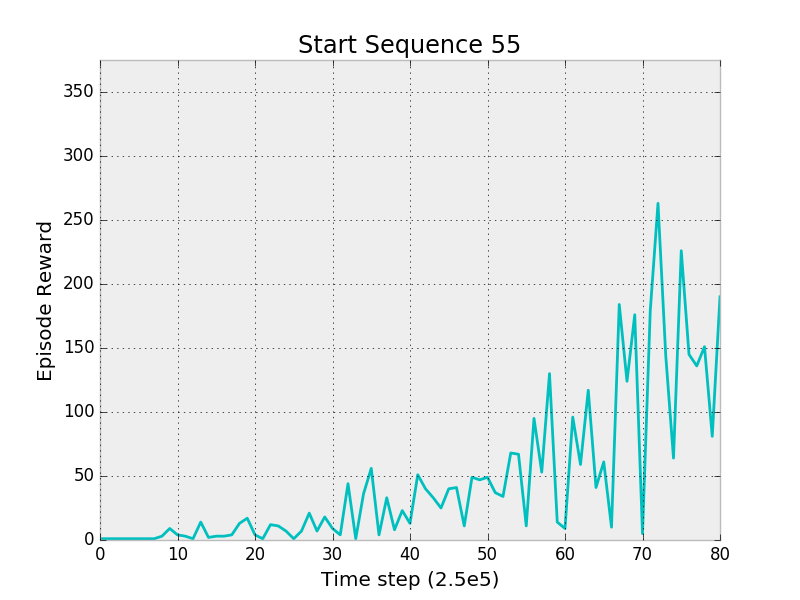}}
    \subfloat[][]{
    	\includegraphics[width=0.25\linewidth]{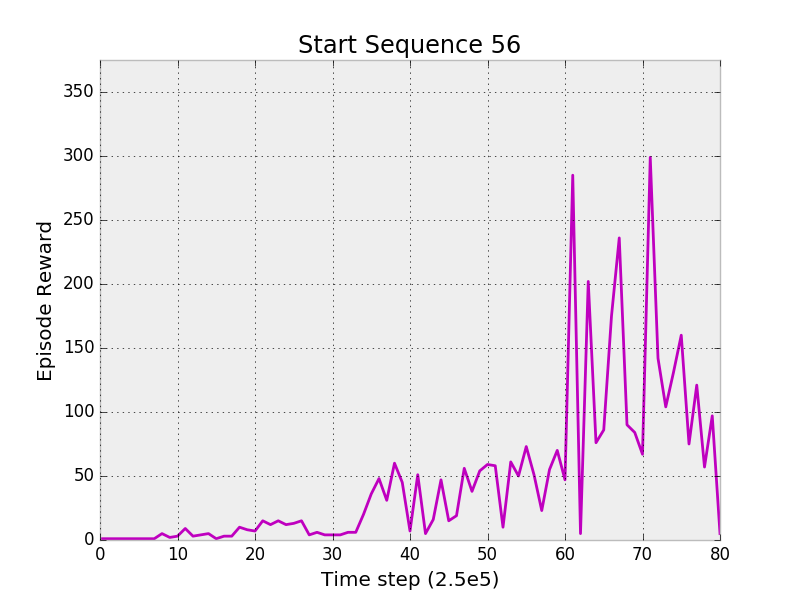}} \\
	\subfloat[][]{
    	\includegraphics[width=0.25\linewidth]{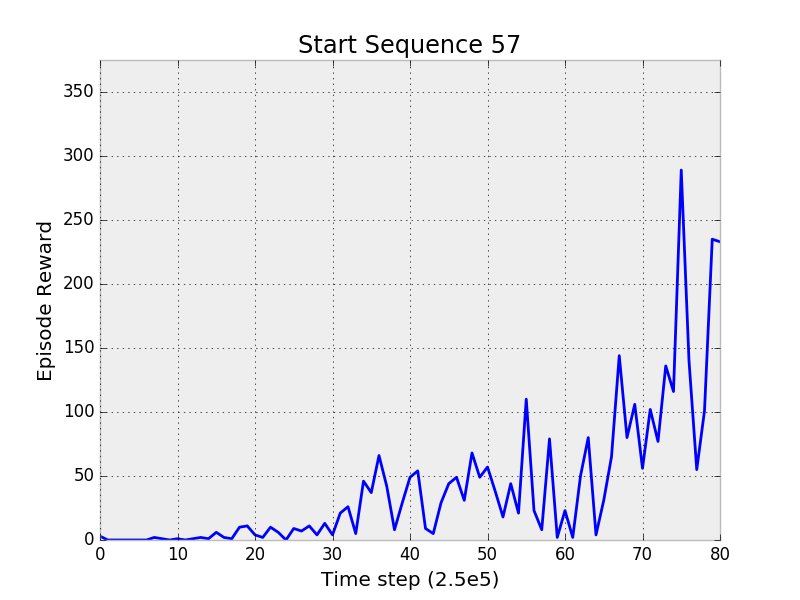}}
    \subfloat[][]{
    	\includegraphics[width=0.25\linewidth]{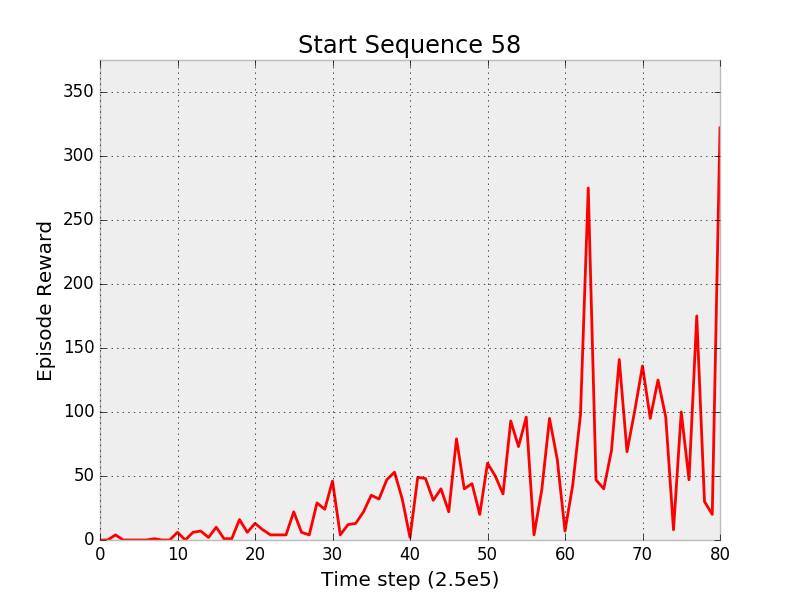}}
	\subfloat[][]{
    	\includegraphics[width=0.25\linewidth]{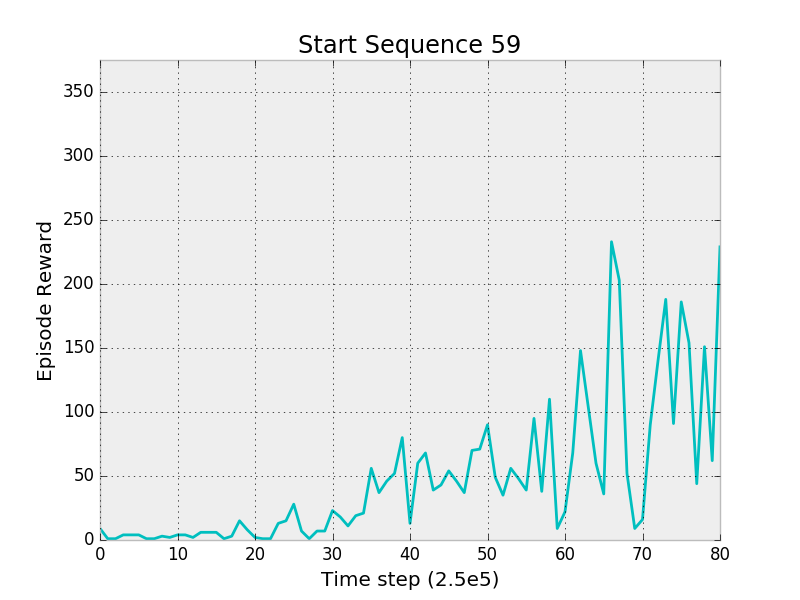}}
    \subfloat[][]{
    	\includegraphics[width=0.25\linewidth]{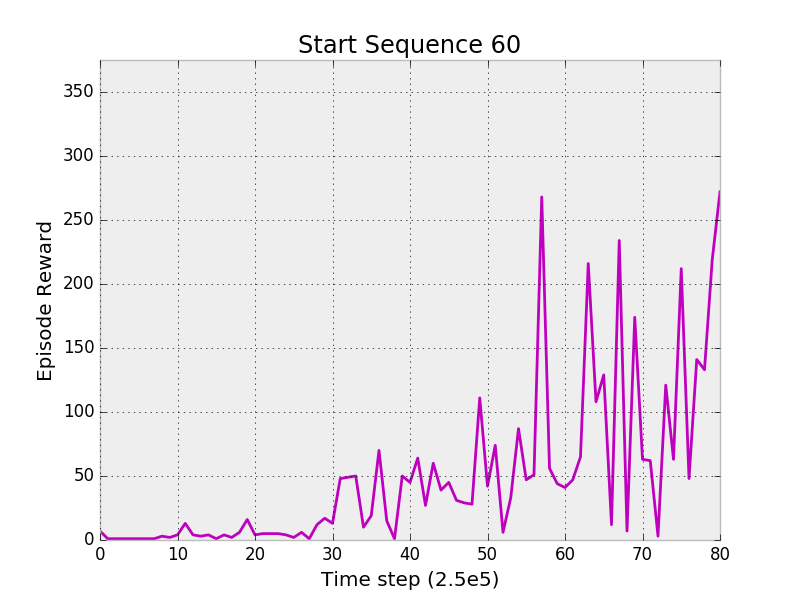}} \\
\caption{Learning curves of the deterministic agent for individual start states/sequences.}
\end{figure}

\begin{figure}[ht]
\ContinuedFloat
\centering
	\subfloat[][]{
    	\includegraphics[width=0.25\linewidth]{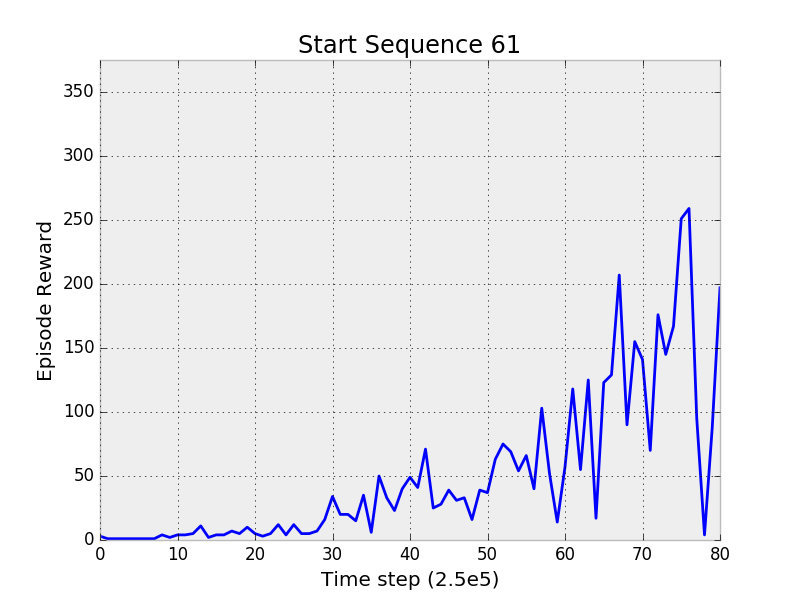}}
    \subfloat[][]{
    	\includegraphics[width=0.25\linewidth]{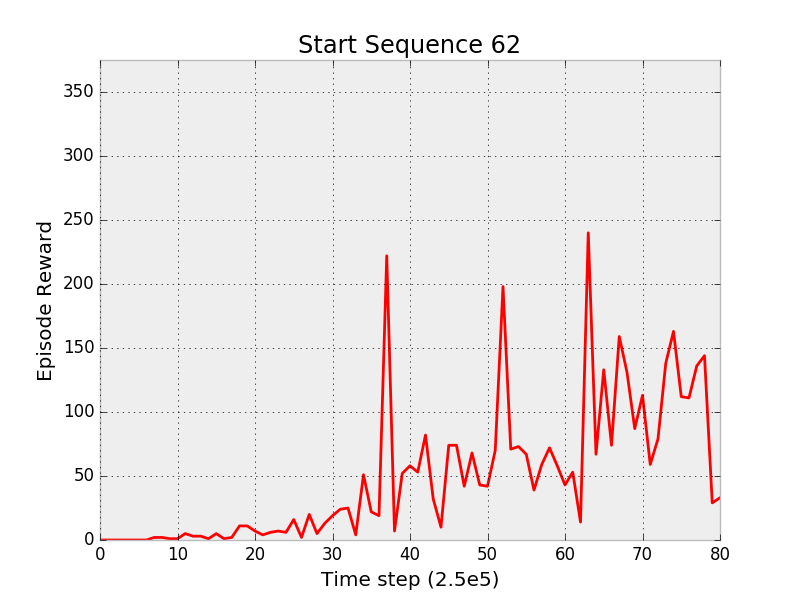}}
	\subfloat[][]{
    	\includegraphics[width=0.25\linewidth]{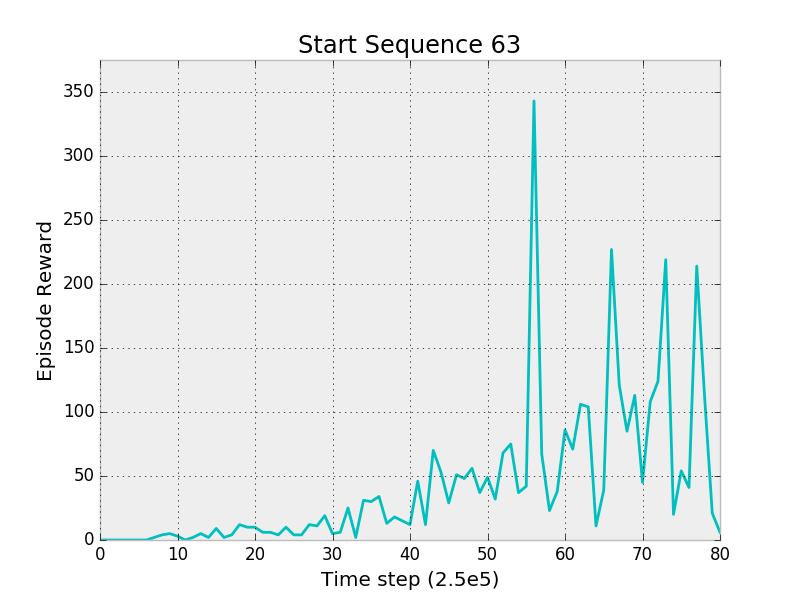}}
    \subfloat[][]{
    	\includegraphics[width=0.25\linewidth]{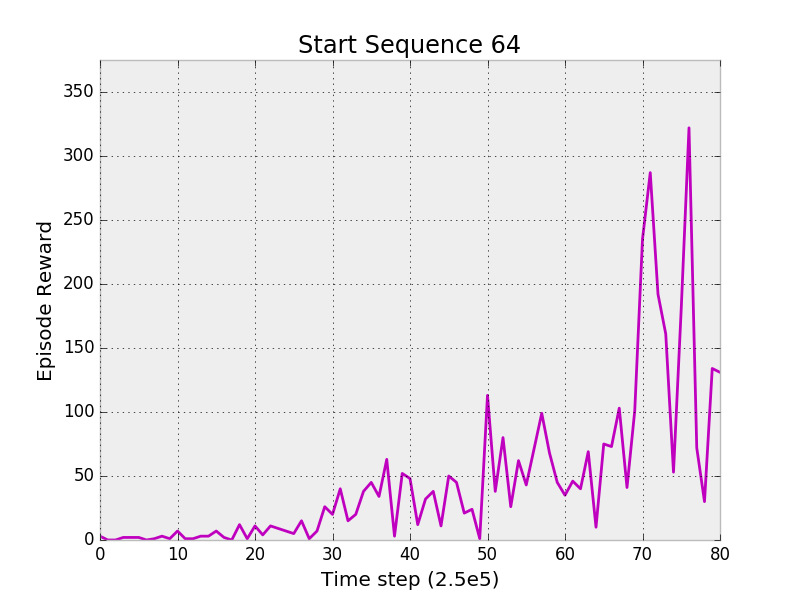}} \\
	\subfloat[][]{
    	\includegraphics[width=0.25\linewidth]{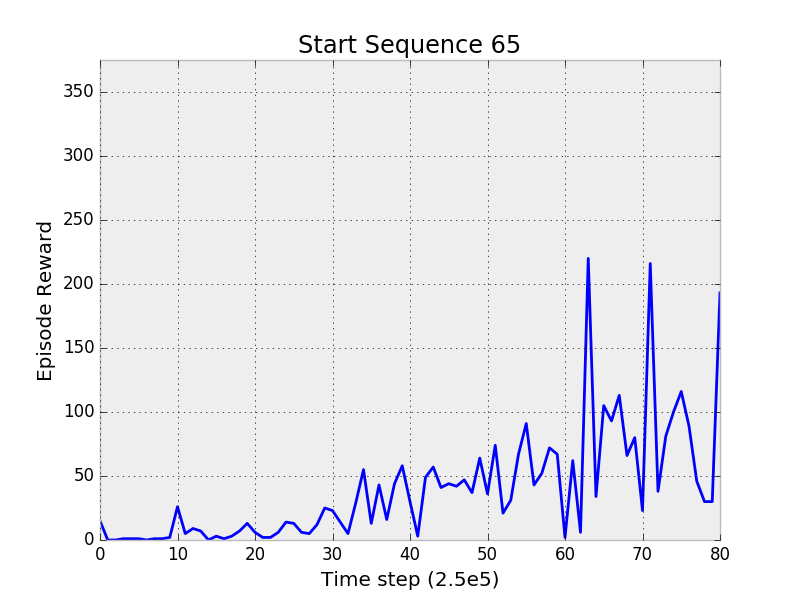}}
    \subfloat[][]{
    	\includegraphics[width=0.25\linewidth]{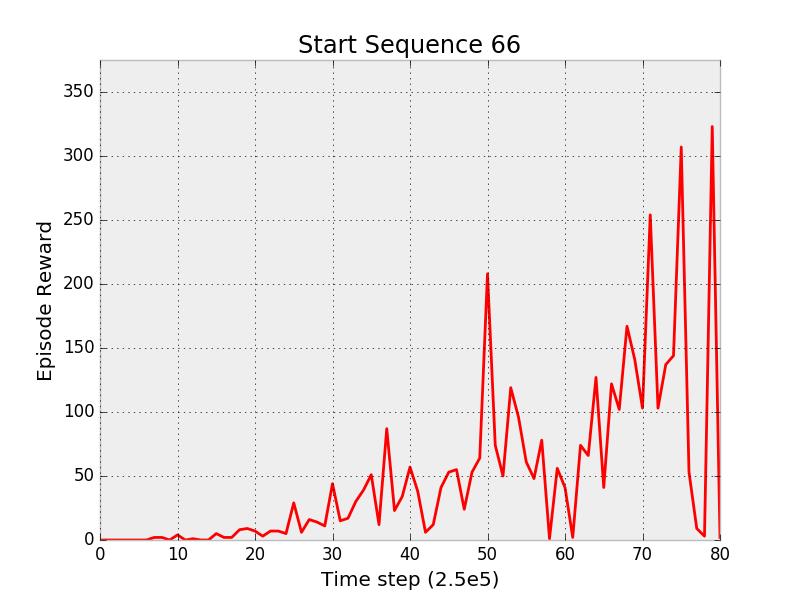}}
	\subfloat[][]{
    	\includegraphics[width=0.25\linewidth]{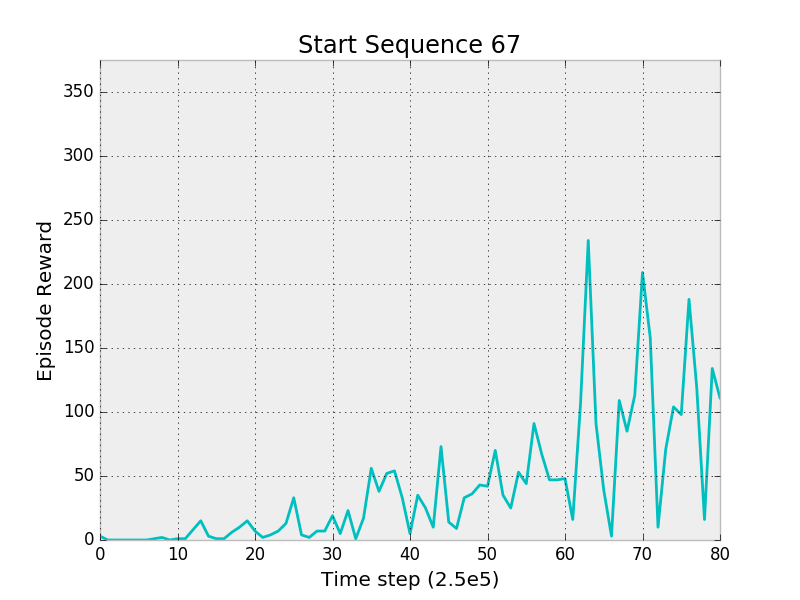}}
    \subfloat[][]{
    	\includegraphics[width=0.25\linewidth]{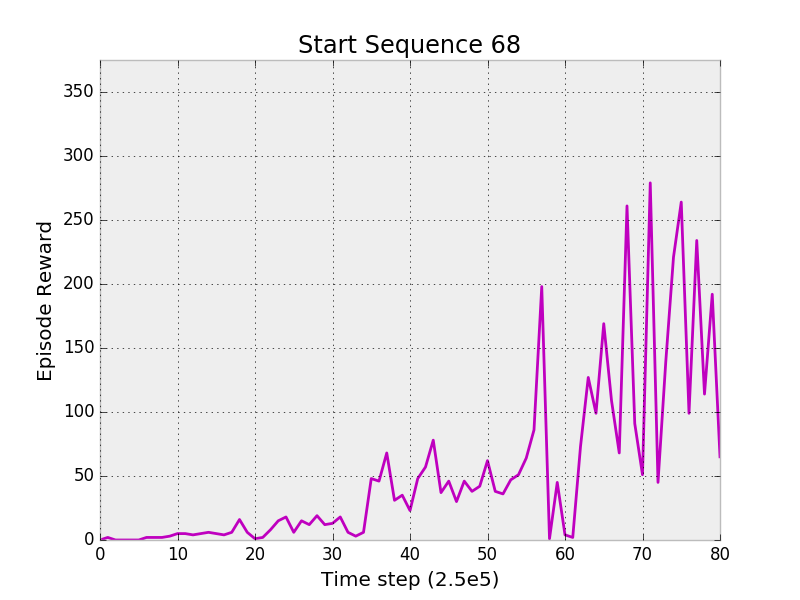}} \\
	\subfloat[][]{
    	\includegraphics[width=0.25\linewidth]{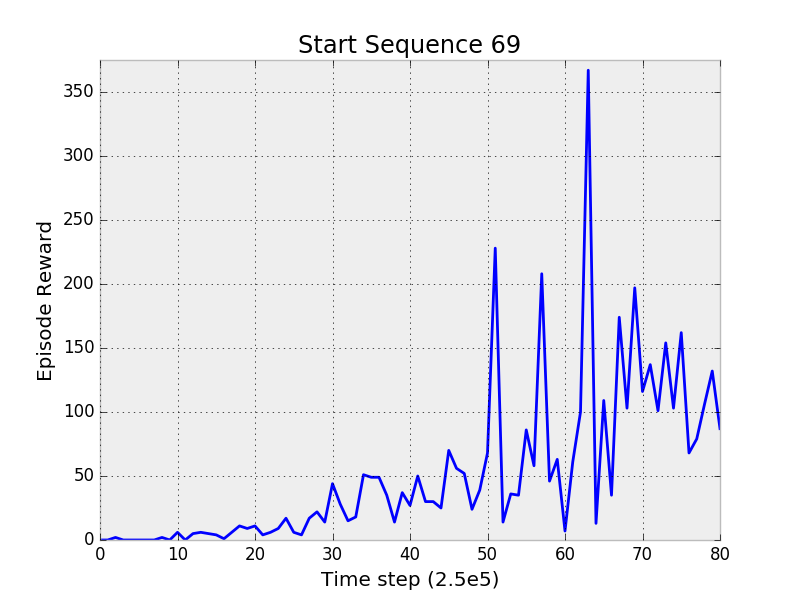}}
    \subfloat[][]{
    	\includegraphics[width=0.25\linewidth]{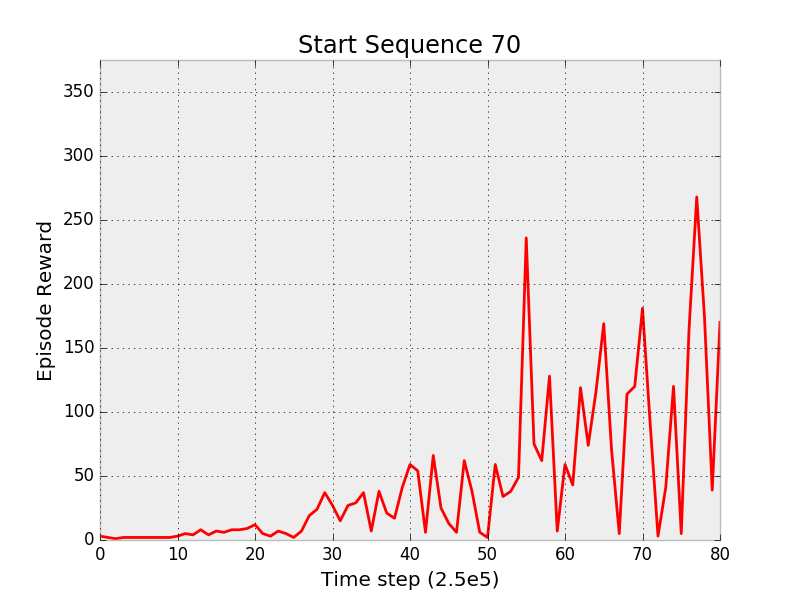}}
	\subfloat[][]{
    	\includegraphics[width=0.25\linewidth]{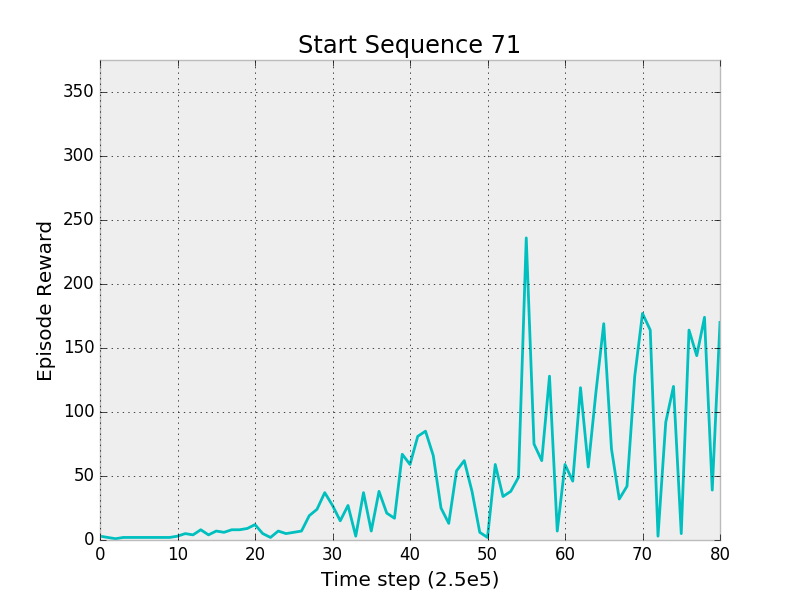}}
    \subfloat[][]{
    	\includegraphics[width=0.25\linewidth]{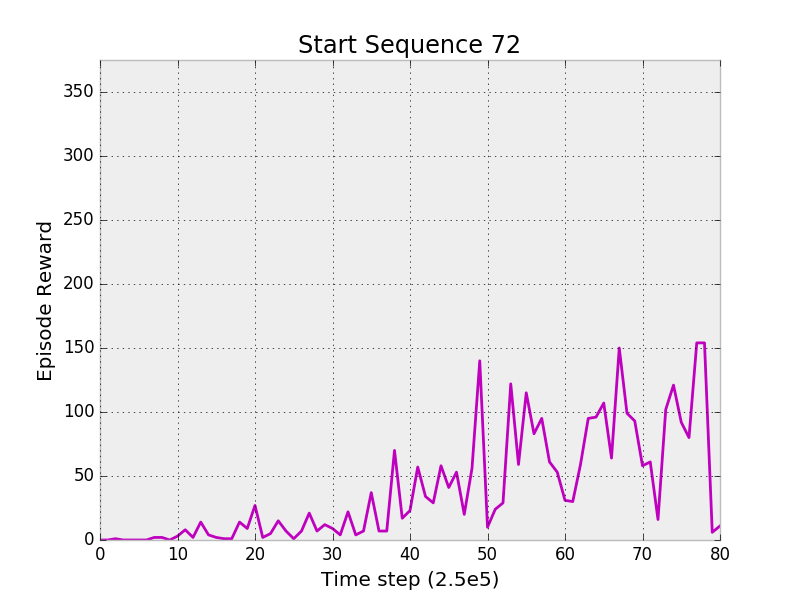}} \\
	\subfloat[][]{
    	\includegraphics[width=0.25\linewidth]{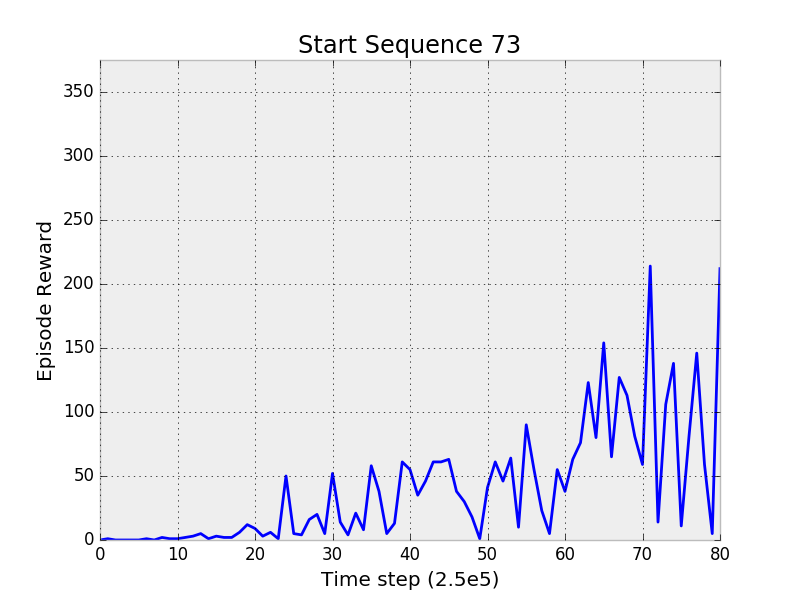}}
    \subfloat[][]{
    	\includegraphics[width=0.25\linewidth]{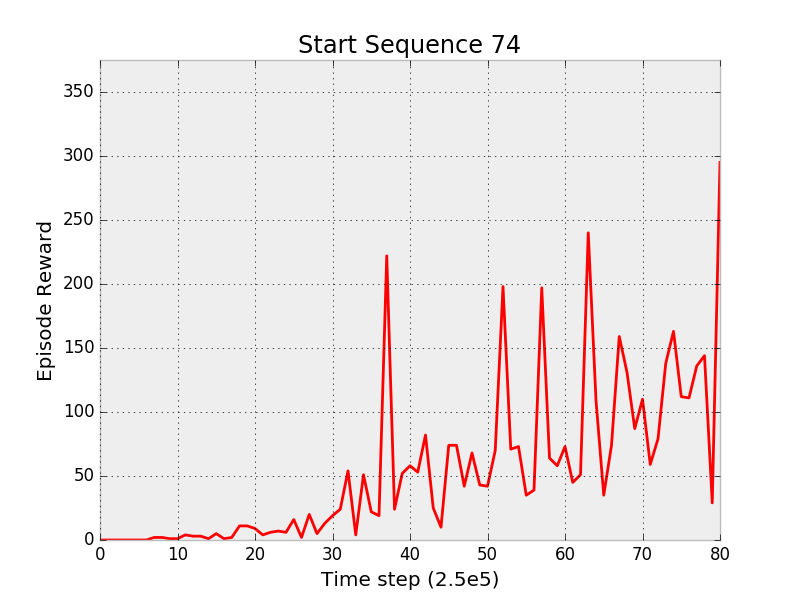}}
	\subfloat[][]{
    	\includegraphics[width=0.25\linewidth]{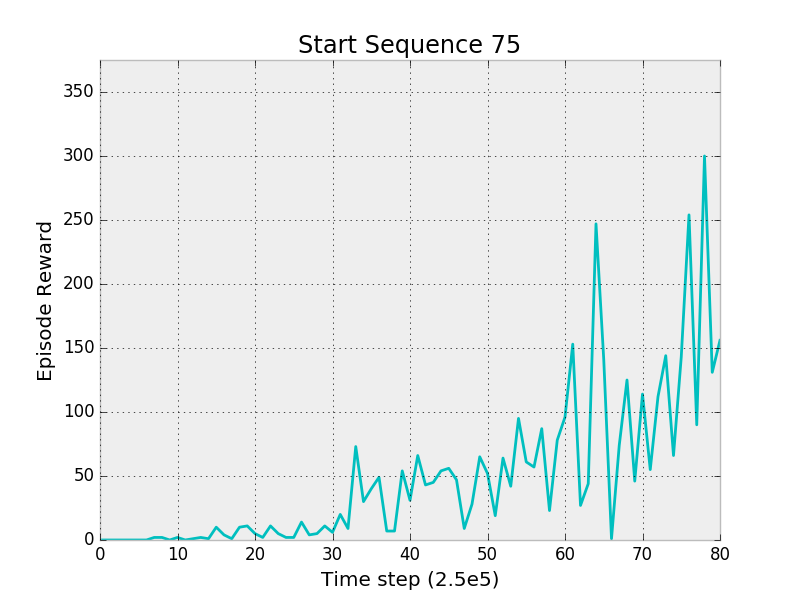}}
    \subfloat[][]{
    	\includegraphics[width=0.25\linewidth]{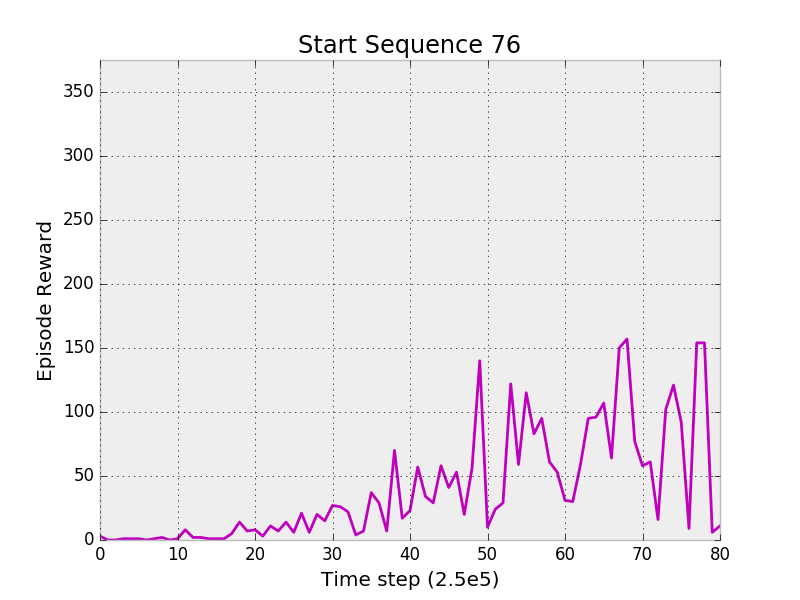}} \\
	\subfloat[][]{
    	\includegraphics[width=0.25\linewidth]{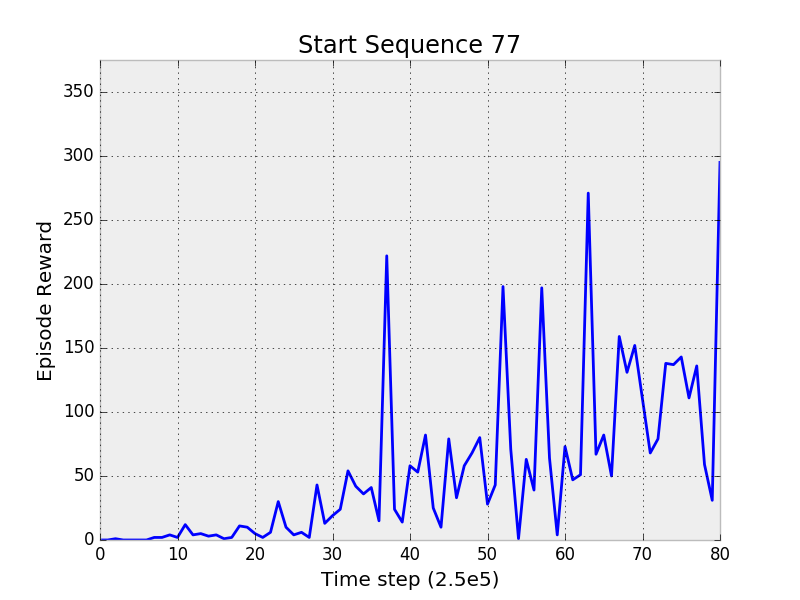}}
    \subfloat[][]{
    	\includegraphics[width=0.25\linewidth]{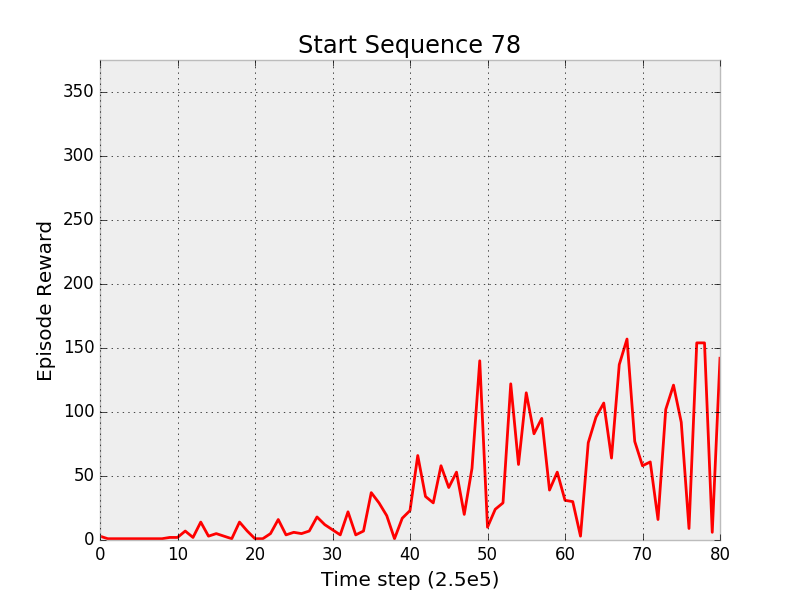}}
	\subfloat[][]{
    	\includegraphics[width=0.25\linewidth]{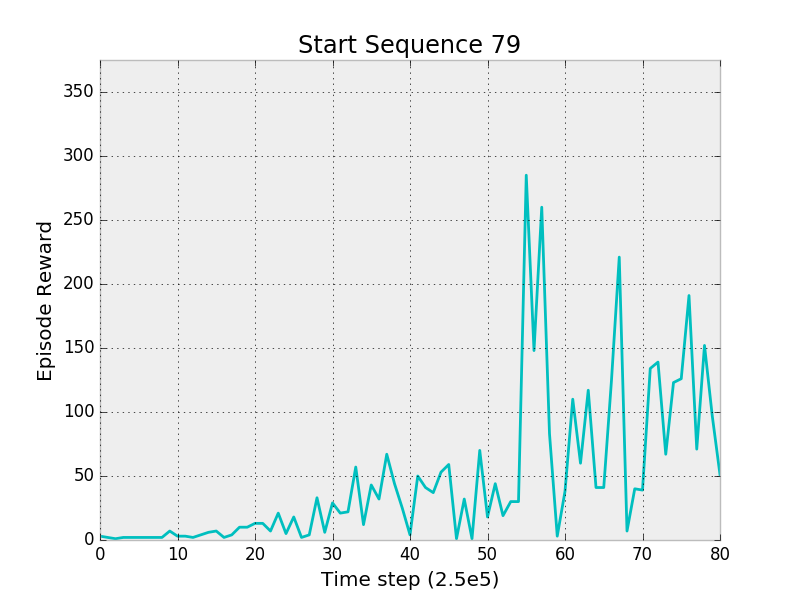}}
    \subfloat[][]{
    	\includegraphics[width=0.25\linewidth]{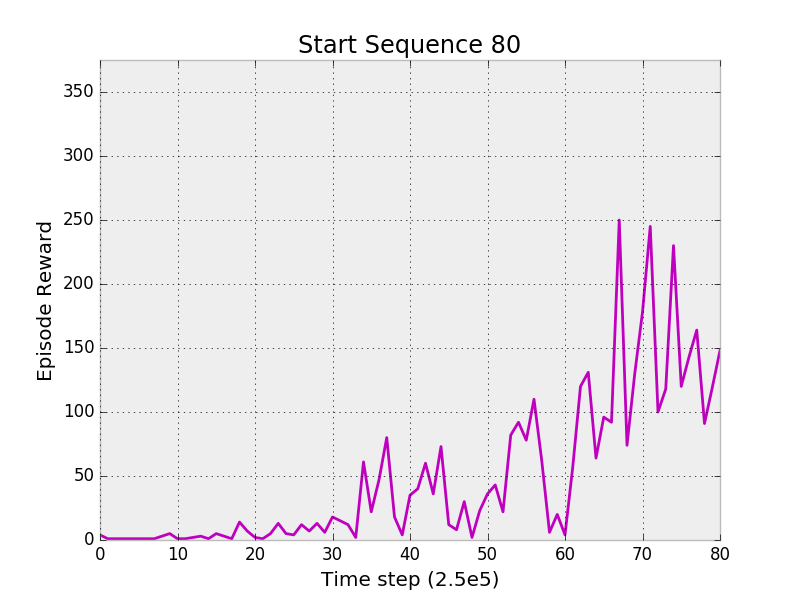}} \\
\caption{Learning curves of the deterministic agent for individual start states/sequences.}
\end{figure}

\begin{figure}[t!]
\ContinuedFloat
\centering
	\subfloat[][]{
    	\includegraphics[width=0.25\linewidth]{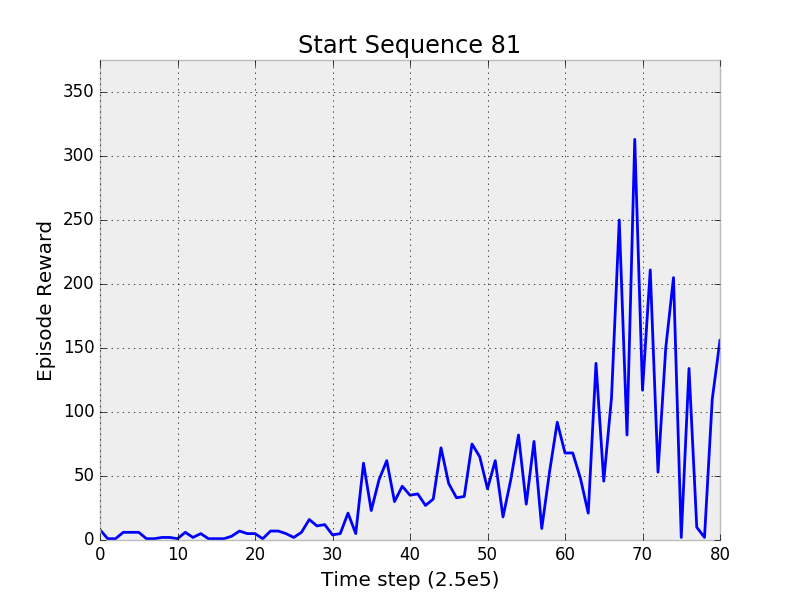}}
    \subfloat[][]{
    	\includegraphics[width=0.25\linewidth]{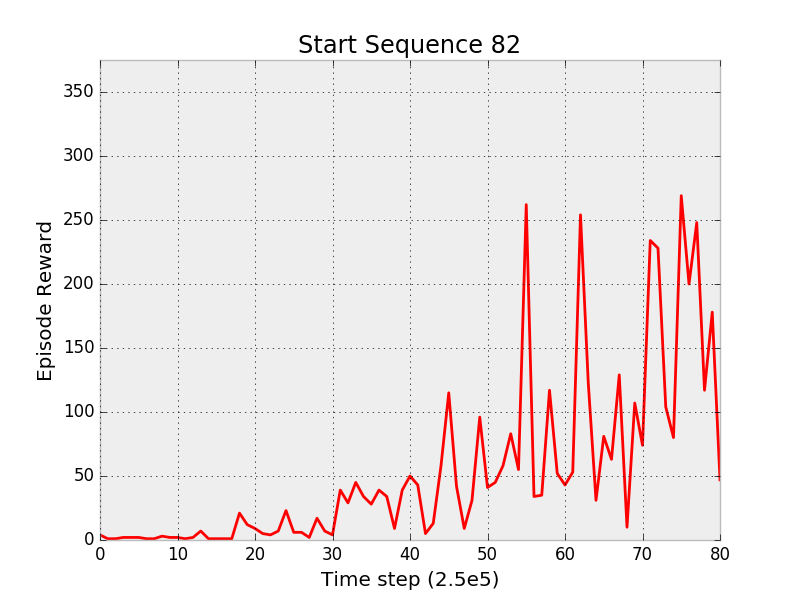}}
	\subfloat[][]{
    	\includegraphics[width=0.25\linewidth]{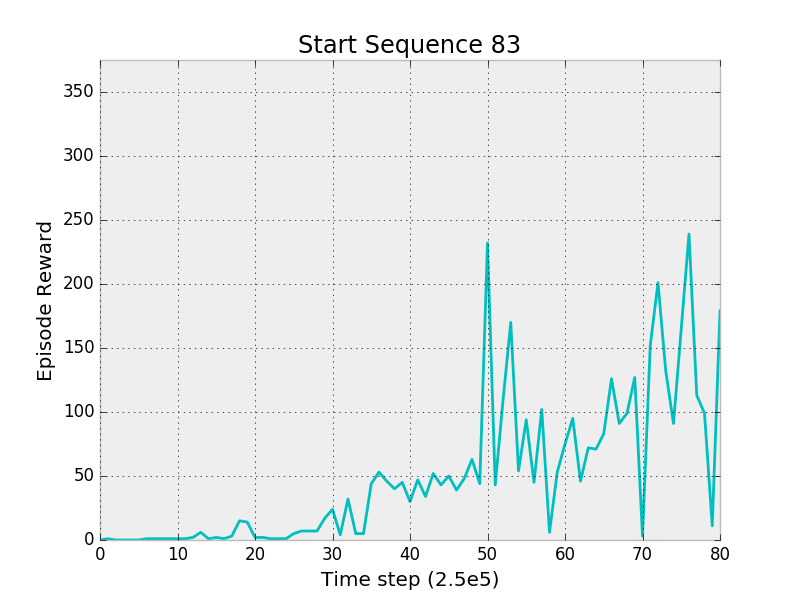}}
    \subfloat[][]{
    	\includegraphics[width=0.25\linewidth]{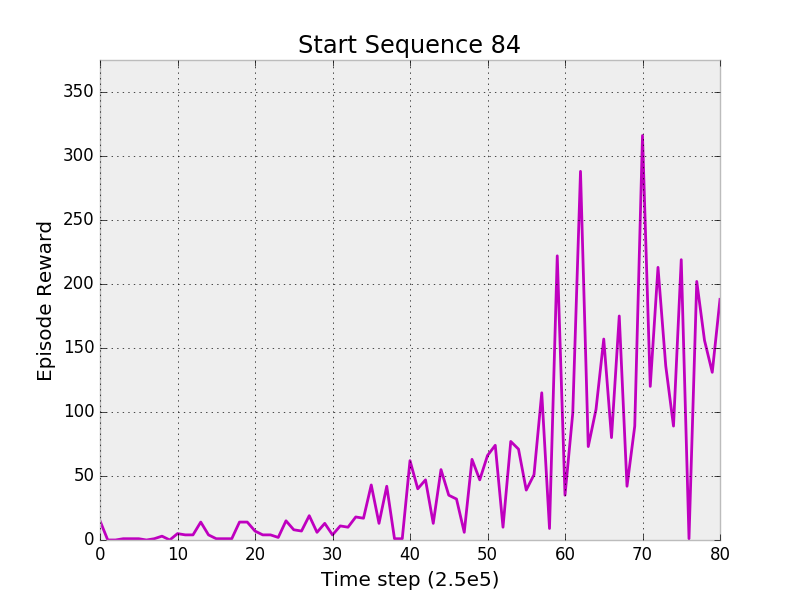}} \\
	\subfloat[][]{
    	\includegraphics[width=0.25\linewidth]{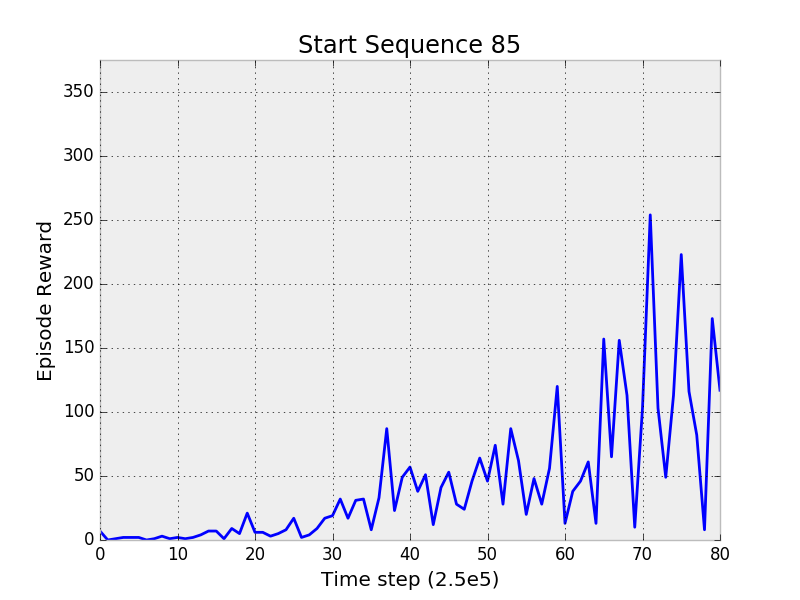}}
    \subfloat[][]{
    	\includegraphics[width=0.25\linewidth]{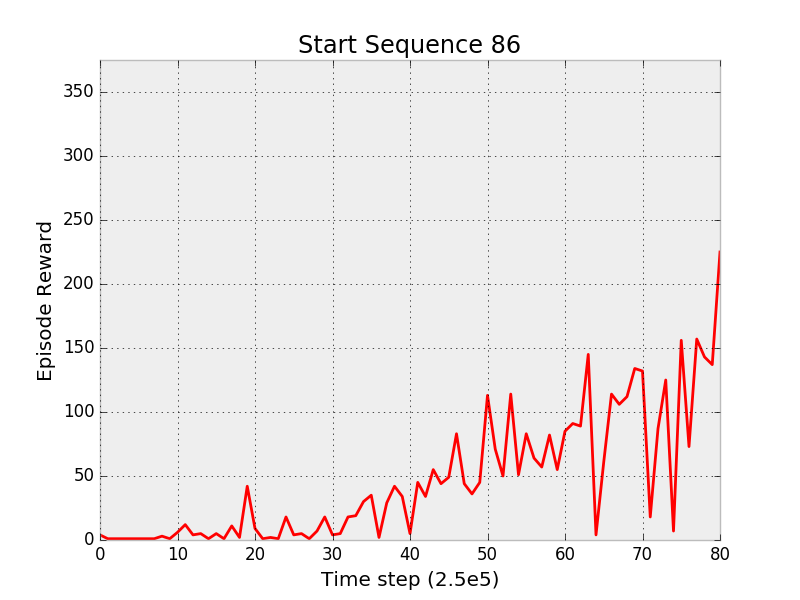}}
	\subfloat[][]{
    	\includegraphics[width=0.25\linewidth]{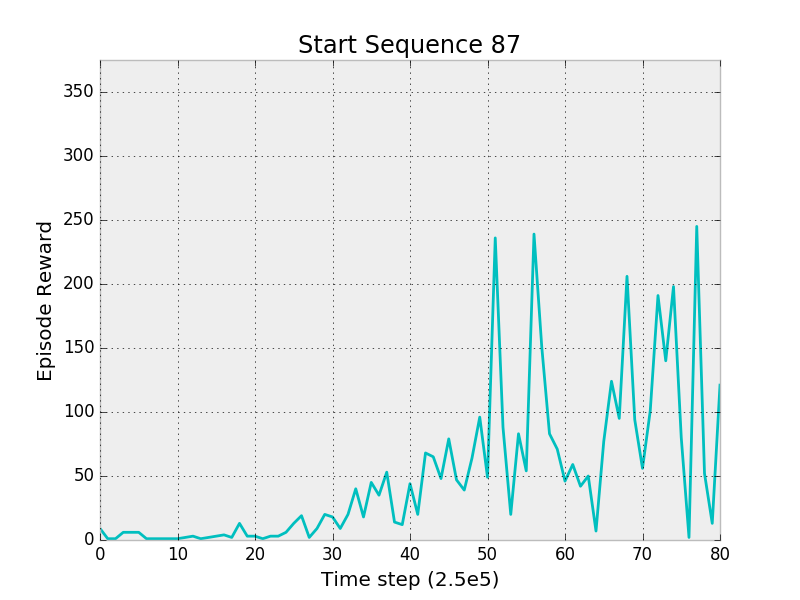}}
    \subfloat[][]{
    	\includegraphics[width=0.25\linewidth]{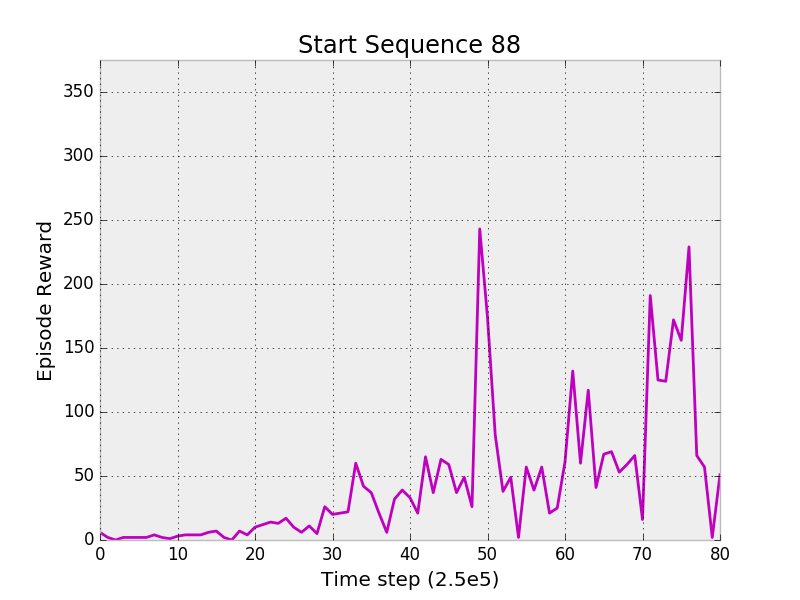}} \\
	\subfloat[][]{
    	\includegraphics[width=0.25\linewidth]{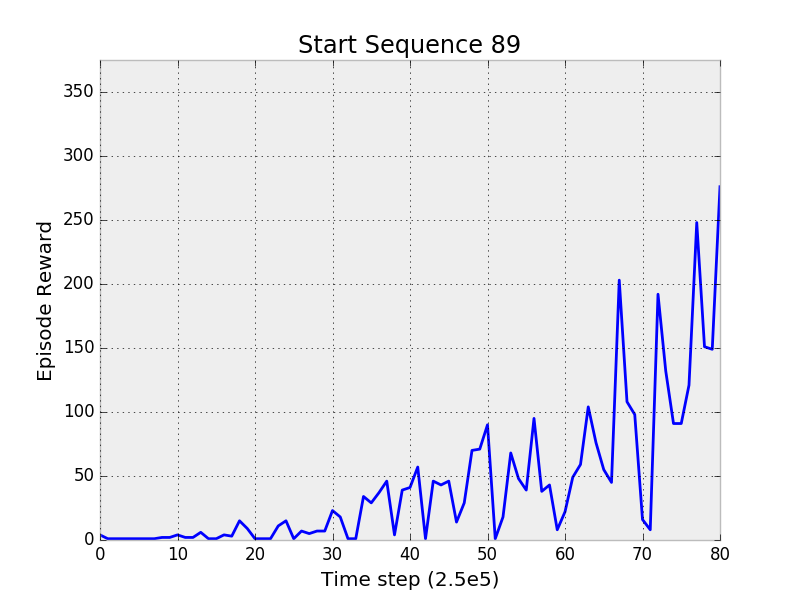}}
    \subfloat[][]{
    	\includegraphics[width=0.25\linewidth]{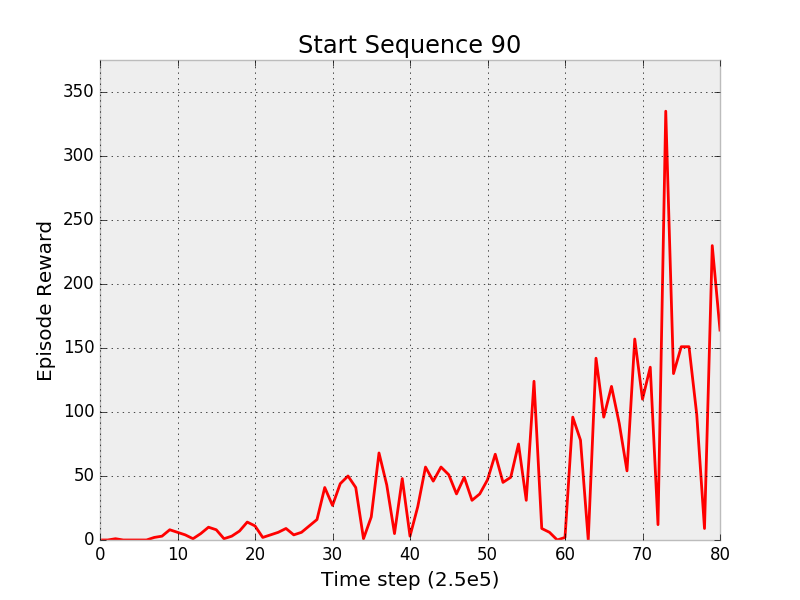}}
	\subfloat[][]{
    	\includegraphics[width=0.25\linewidth]{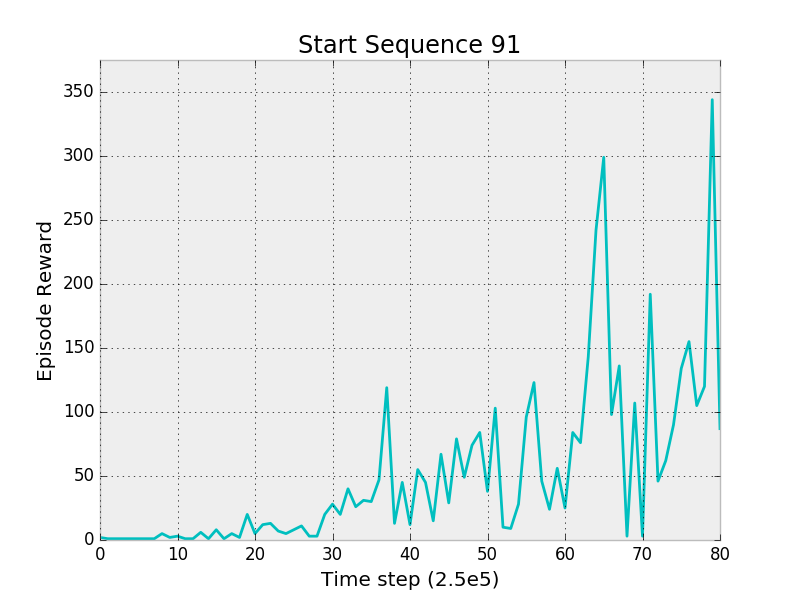}}
    \subfloat[][]{
    	\includegraphics[width=0.25\linewidth]{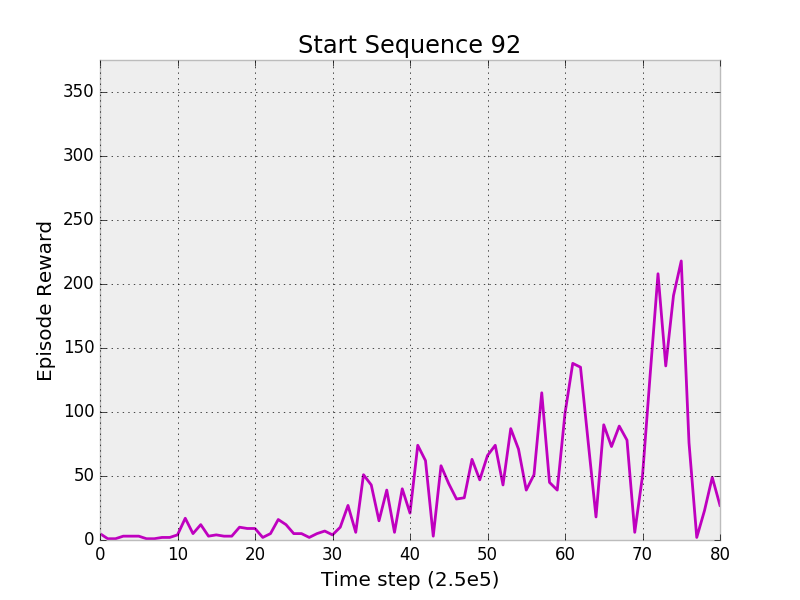}} \\
	\subfloat[][]{
    	\includegraphics[width=0.25\linewidth]{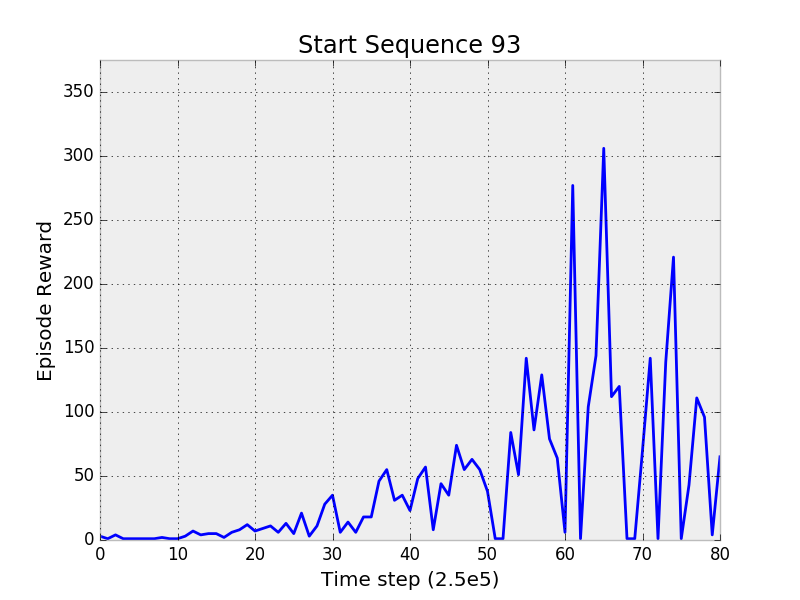}}
    \subfloat[][]{
    	\includegraphics[width=0.25\linewidth]{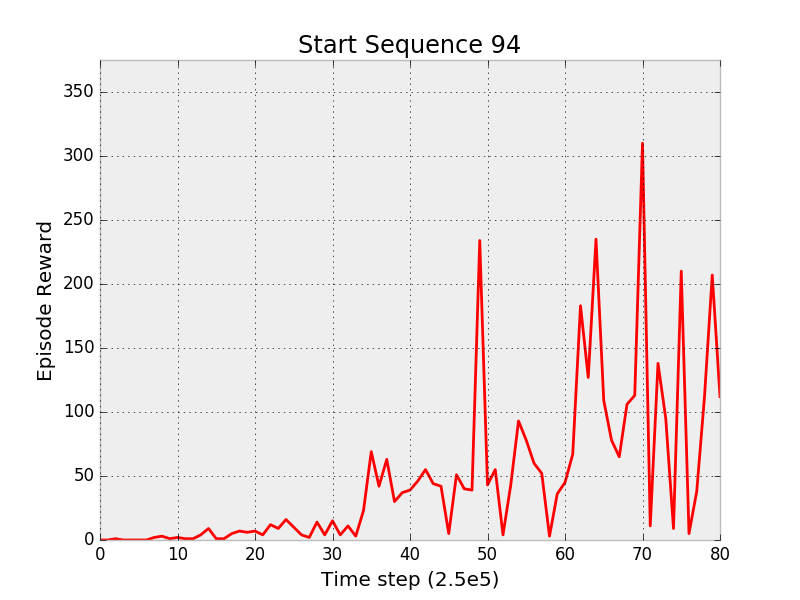}}
	\subfloat[][]{
    	\includegraphics[width=0.25\linewidth]{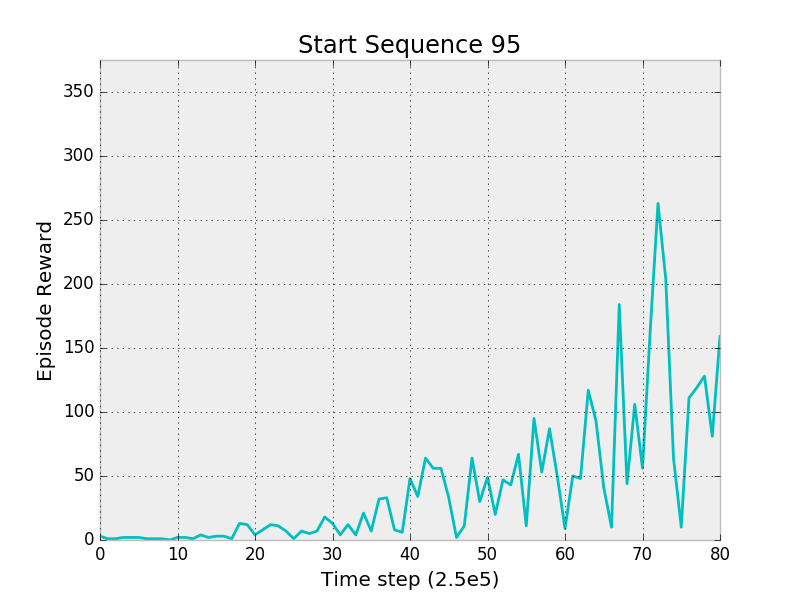}}
    \subfloat[][]{
    	\includegraphics[width=0.25\linewidth]{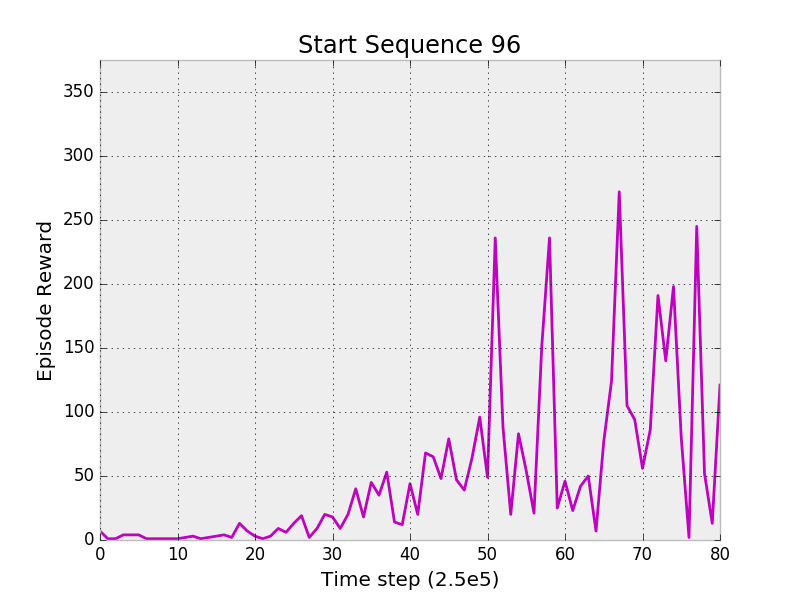}} \\
	\subfloat[][]{
    	\includegraphics[width=0.25\linewidth]{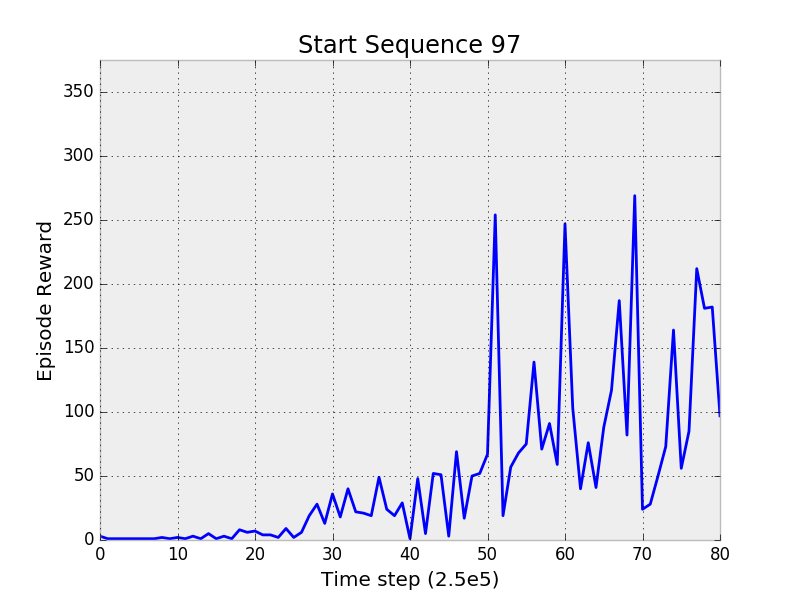}}
    \subfloat[][]{
    	\includegraphics[width=0.25\linewidth]{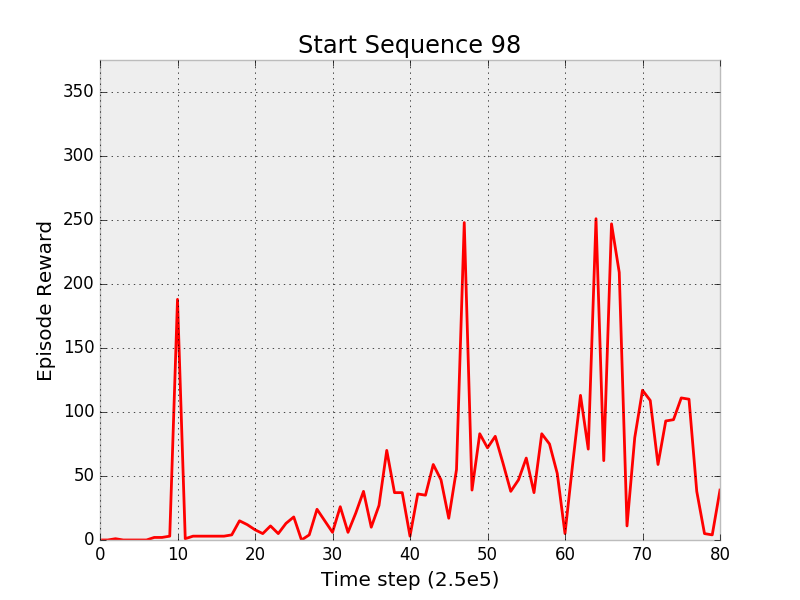}}
	\subfloat[][]{
    	\includegraphics[width=0.25\linewidth]{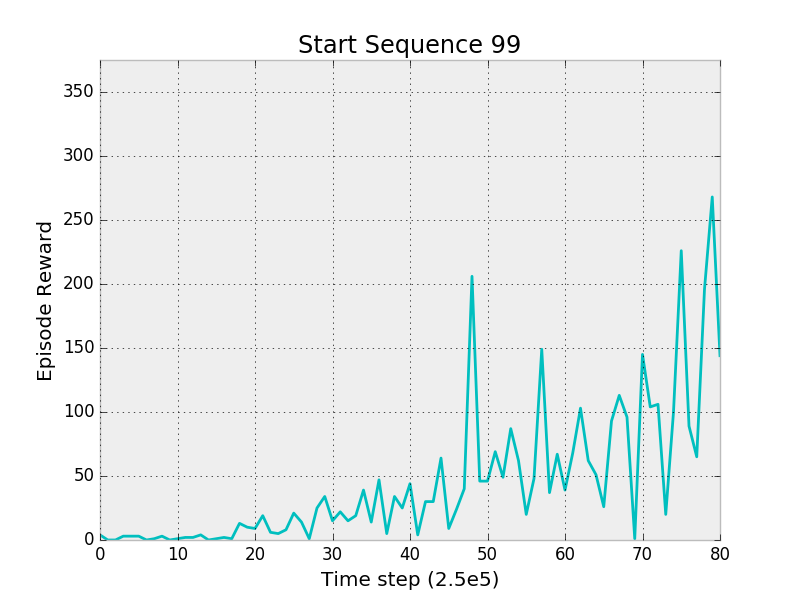}}
    \subfloat[][]{
    	\includegraphics[width=0.25\linewidth]{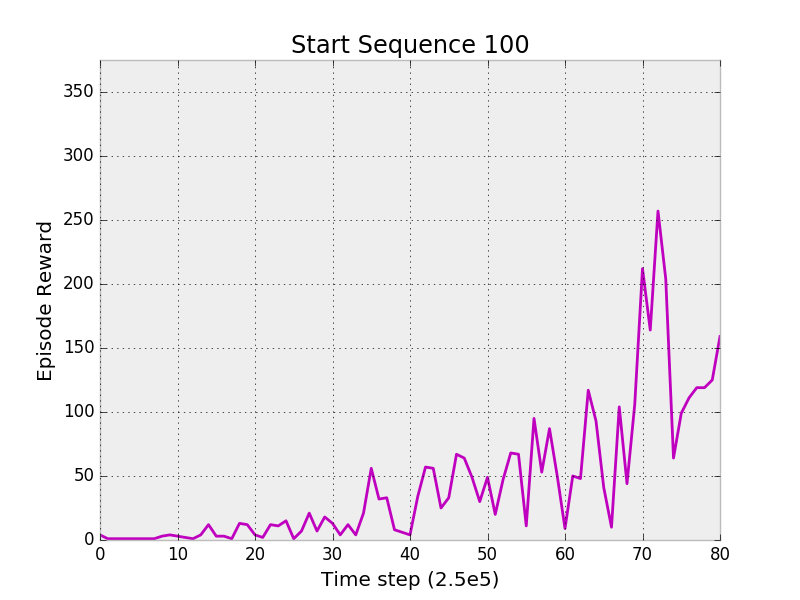}} \\
\caption{Learning curves of the deterministic agent for individual start states/sequences.}
\end{figure}

\end{document}